\documentclass[final,authoryear]{elsarticle}

\usepackage{ecrc}

\volume{00}

\firstpage{1}

\journalname{iScience}

\runauth{}

\jid{procs}

\jnltitlelogo{iScience}

\CopyrightLine{2011}{Published by Elsevier Ltd.}

\usepackage{amssymb}
\usepackage{graphicx}
\usepackage{amsmath}
\usepackage{amssymb}
\usepackage{array}
\usepackage{paralist}
\usepackage{caption}
\RequirePackage{amsthm,amsmath,amssymb}
\RequirePackage{hyperref}

\usepackage{booktabs}
\usepackage{algorithm}
\usepackage{tabularx}
\usepackage{todonotes}

\usepackage{amsthm}
\usepackage{mathtools}

\usepackage{subfigure}
\usepackage{url}
\usepackage{enumerate}
\usepackage{enumitem}

\usepackage{color}
\usepackage{xcolor}

\usepackage{url}
\usepackage{scrextend}

\usepackage{cleveref}

\usepackage[
  separate-uncertainty = true,
  multi-part-units = repeat,
  detect-weight = true 
]{siunitx}
\usepackage{booktabs}
\usepackage{etoolbox,siunitx}
\robustify\bfseries
\usepackage[symbol]{footmisc}

\theoremstyle{plain}

\usepackage{commath}

\usepackage{subfigure}
\usepackage{enumerate}
\usepackage{enumitem}

\usepackage{algpseudocode}

\newcommand{\STAB}[1]{\begin{tabular}{@{}c@{}}#1\end{tabular}}
\usepackage{array,multirow,graphicx}

\newcolumntype{P}[1]{>{\centering\arraybackslash}p{#1}}

\usepackage[switch]{lineno} 
\usepackage{xcolor}
\DeclareRobustCommand{\rchi}{{\mathpalette\irchi\relax}}
\newcommand{\irchi}[2]{\raisebox{\depth}{$#1\chi$}}

\usepackage[figuresright]{rotating}

\begin{document}

\begin{frontmatter}



\dochead{}
%% e.g. \dochead{17th International Conference on Dynamical Processes in Excitepd States of Solids}

\title{CX-ToM: Counterfactual Explanations with Theory-of-Mind for Enhancing Human Trust in Image Recognition Models}

\author[1]{Arjun~R.~Akula*}
\author[1]{Keze~Wang*}
\author{Changsong~Liu}
\author[2]{Sari~Saba-Sadiya}
\author[1]{Hongjing~Lu}
\author[3]{Sinisa~Todorovic}
\author[2]{Joyce~Chai}
\author[4]{Song-Chun~Zhu}
\address[1]{Department of Statistics, UCLA, CA, 90024, USA}
\address[2]{Department of Computer Science, University of Michigan, Ann Arbor, MI, 48109, USA}
\address[3]{Department of Computer Science, Oregon State University, OR, 97331, USA}
\address[4]{Beijing Institute for General AI (BIGAI); Tsinghua University; Peking University, 100871, China}
\footnote{Correspondence (Lead Contact): aakula@ucla.edu}

% \author{Arjun~R.~Akula*, %~\IEEEmembership{Member,~IEEE,}        
%         Keze~Wang*, %~\IEEEmembership{Member,~IEEE,}
%         Changsong~Liu, %~\IEEEmembership{Fellow,~IEEE,}
%         Sari~Saba-Sadiya, %~\IEEEmembership{Member,~IEEE,}
%         Hongjing~Lu, %~\IEEEmembership{Member,~IEEE,}
%         Sinisa~Todorovic, %~\IEEEmembership{Fellow,~IEEE,}
%         Joyce~Chai, %~\IEEEmembership{Fellow,~IEEE,}
%         and~Song-Chun~Zhu%~\IEEEmembership{Fellow,~IEEE}% <-this % stops a space
% \footnote{asa}
% %\IEEEcompsocitepmizethanks{\IEEEcompsocthanksitem Arjun R. Akula: University of California, Los Angeles (UCLA), email: aakula@ucla.edu;\\ Keze Wang: UCLA, kezewang@gmail.com; \\ Changsong Liu: liucs.msu@gmail.com; \\ Sari~Saba-Sadiya: Michigan State University, sadiyasa@cse.msu.edu;\\ Hongjing Lu: UCLA, hongjing@ucla.edu; \\ Sinisa Todorovic: Oregon State University, sinisa@oregonstate.edu;\\ Joyce Chai: University of Michigan, Ann Arbor, jchai@cse.msu.edu;\\ Song-Chun Zhu: Beijing Institute for General Artificial Intelligence (BIGAI), Tsinghua University, Peking University, s.c.zhu@pku.edu.cn\protect}}
% }

% \IEEEcompsocthanksitem * denotes equal contribution.\protect}% <-this % stops an unwanted space
% \thanks{Manuscript received Sep, 2020; revised Aug, 2021.}}

% \markboth{iScience Cell Press 2021}%
% {Shell \MakeLowercase{}}

\address{}

\begin{abstract}
We propose \textit{CX-ToM}, short for counterfactual explanations with theory-of-mind, a new explainable AI (XAI) framework for explaining decisions made by a deep convolutional neural network (CNN). In contrast to the current methods in XAI that generate explanations as a single shot response, we pose explanation as an iterative communication process, i.e. dialog, between the machine and human user. More concretely, our CX-ToM framework generates a sequence of explanations in a dialog by mediating the differences between the minds of the machine and human user. To do this, we use Theory of Mind (ToM) which helps us in explicitly modeling human's intention, machine's mind as inferred by the human as well as human's mind as inferred by the machine. Moreover, most state-of-the-art XAI frameworks provide attention (or heat map) based explanations. In our work, we show that these attention-based explanations are not sufficient for increasing human trust in the underlying CNN model. In CX-ToM, we instead use counterfactual explanations called \textit{fault-lines} which we define as follows: given an input image $I$ for which a CNN classification model $M$ predicts class $c_{pred}$, a fault-line identifies the minimal semantic-level features (e.g., \textit{stripes} on zebra), referred to as explainable concepts, that need to be added to or deleted from $I$ in order to alter the classification category of $I$ by $M$ to another specified class $c_{alt}$. Extensive experiments verify our hypotheses, demonstrating that our CX-ToM significantly outperforms the state-of-the-art XAI models.
\end{abstract}


\end{frontmatter}

% \linenumbers

\section{Introduction}\label{sec:introduction}

Intelligence (AI) systems are becoming increasingly ubiquitous from low risk environments such as movie recommendation systems and chatbots to high-risk environments such as medical-diagnosis and treatment, self-driving cars, drones and military applications~\citep{chancey2015role,gulshan2016development,lyons2017certifiable,DBLP:journals/corr/MnihKSGAWR13,gupta2012novel,pulijala2013web,dasgupta2014towards,agarwal2017automatic,palakurthi2015classification,akula2021measuring,akula2021crossvqa,akula2021mind,akula2021robust}. In particular, AI systems built using black box machine learning (ML) models -- such as deep neural networks and large ensembles~\citep{lipton2016mythos,ribeiro2016should,miller2017explanation,yang2018,sundararajan2017axiomatic,ramprasaath2016grad,zeiler2014visualizing,smilkov2017smoothgrad,kim2014bayesian,akula2013novel,akula2020words,akula2015novel} -- perform remarkably well on a broad range of tasks and are gaining widespread adoption. However, understanding and developing human trust in these systems remains a significant challenge as they cannot explain why they reached a specific recommendation or a decision. This is especially problematic in high-risk environments such as banking, healthcare, and insurance, where AI decisions can have significant consequences.

In light of aforementioned issues, eXplainable Artificial Intelligence (XAI) has become an active area of interest in research community and industry. XAI models, through explanations, aim at making the underlying inference mechanism of AI systems transparent and interpretable to expert users (system developers) and non-expert users (end-users)~\citep{lipton2016mythos,ribeiro2016should,hoffman17explanation}.  In this work, we focus mainly on increasing justified human trust (JT) in a deep convolutional neural network (CNN), through explanations~\citep{hoffman2018metrics,akula2019x,akula2019explainable}. Justified trust is computed based on human judgments of CNN model's prediction (more details on this are described in Section \ref{sec:1.2}). Despite an increasing amount of work on XAI~\citep{smilkov2017smoothgrad,sundararajan2017axiomatic,zeiler2014visualizing,kim2014bayesian,zhang2018interpretable,r2019natural}, providing explanations that can increase justified human trust remains an important research problem~\citep{DBLP:journals/corr/abs-1902-10186}. 

\begin{figure*}[h]
\centering
  \includegraphics[width=0.65\linewidth]{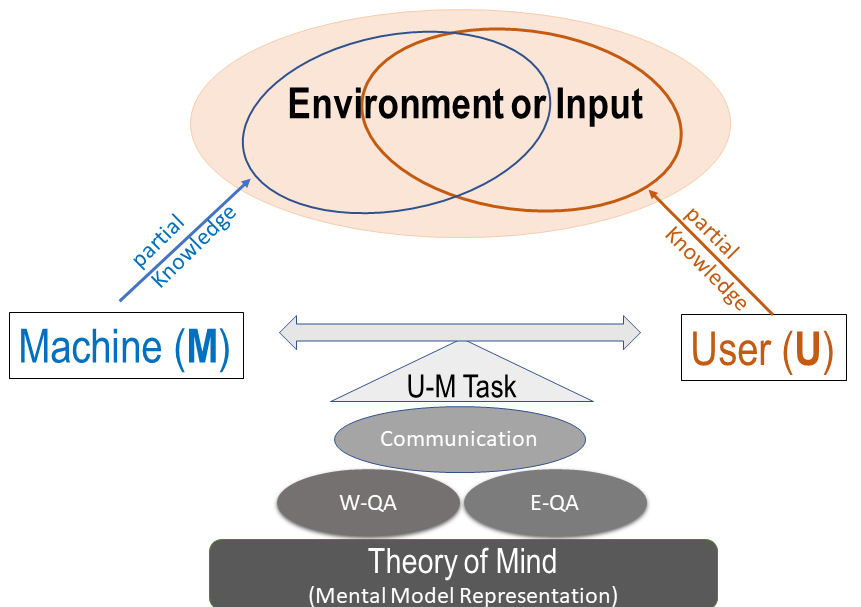}
  \caption{\textbf{CX-ToM}: Our interactive and collaborative XAI framework based on the Theory of Mind. The interaction is conducted through a dialog where the user poses questions about facts in the environment (W-QA) and explanation seeking questions (E-QA).}~\label{fig:2agents}
\end{figure*}

Our work is motivated by the following two key observations:
\begin{enumerate}
    \item \textbf{Attention is not a Good Explanation:} Previous studies have shown that trust is closely and positively correlated to the level of how much human users understand the AI  system --- {\em understandability} --- and how accurately they can predict the system's performance on a given task --- {\em predictability}~\citep{hoffman17explanation,lipton2016mythos,hoffman2018metrics,miller2017explanation}. Hence, there has been a growing interest in developing explainable AI systems (XAI) aimed at increasing understandability and predictability by providing explanations about the system's predictions to human users~\citep{lipton2016mythos,ribeiro2016should,miller2017explanation,yang2018}. Current works on XAI generate explanations about their performance in terms of, e.g., feature visualization and attention maps~\citep{sundararajan2017axiomatic,ramprasaath2016grad,zeiler2014visualizing,smilkov2017smoothgrad,kim2014bayesian,zhang2018interpretable}. However, solely generating explanations, regardless of their type (visualization or attention maps) and utility, {\em is not sufficient} for increasing understandability and predictability~\citep{DBLP:journals/corr/abs-1902-10186}. We verify this in our experiments. 

\item \textbf{Explanation is an Interactive Communication Process:} We believe that an effective explanation cannot be one shot and involves iterative process of communication between the human and the machine. The context of such interaction plays an important role in determining the utility of the follow-up explanations~\citep{clark1989contributing}. As humans can easily be overwhelmed with too many or too detailed explanations, interactive communication process helps in understanding the user and identify user-specific content for explanation. Moreover, cognitive studies~\citep{miller2017explanation} have shown an explanation can only be optimal if it is generated by taking user's perception and belief into account.  %As illustrated in Figure~\ref{fig:2agents}, we developed ToM based explanation framework, called \textbf{X-ToM}, for optimizing an interactive process of providing explanations to the user toward increasing human trust in the underlying AI system. 
\end{enumerate}

Based on the above two key observations, we introduce an interactive explanation framework, \textbf{CX-ToM}. Unlike current XAI methods that model the explanation as a single shot response, in CX-ToM, we pose the explanation generation as an iterative process of communication between the human and the machine. Central to our approach is the use of Theory-of-Mind (ToM)~\citep{devin2016implemented,goldman2012oxford,premack1978does,bara2019} in driving the iterative dialog by taking into account three important aspects at each dialog turn: (a) human's intention (or curiosity); (b) human's understanding of the machine; and (c) machine's understanding of the human user. Specifically, in our framework, the machine and the user are positioned to solve a collaborative task, but the machine's mind ($M$) and the human user's mind ($U$) only have a partial knowledge of the environment (see Figure~\ref{fig:2agents}). Hence, the machine and user need to communicate with each other, using their partial knowledge, otherwise they would not be able to optimally solve the collaborative task. The communication consists of two different types of question-answer (QA) exchanges --- namely, a) Factoid question-answers about the environment (W-QA), where the user asks ``WH''-questions that begin with  \texttt{what}, \texttt{which}, \texttt{where}, and \texttt{how}; and b) Explanation seeking question-answers (E-QA), where the user asks questions that begin with \texttt{why} about the machine's inference. 

%\begin{figure*}[t]
%\centering
%   \includegraphics[width=0.75\linewidth]{intro2.pdf}
%   \caption{\textbf{Fault-Line Explanations in CX-ToM}: Positive fault-line explanation ($\Psi^{+}_{I_1}$) suggests adding \textit{stripes} to the animal in the input image ($I_1$) to alter the model $M$'s prediction from \texttt{Dog} class to \texttt{Thylacine} class, i.e., the concept of \textit{stripedness} is  critical for $M$ to decide between \texttt{Dog} and \texttt{Thylacine} in $I_1$. Similarly, negative fault-line $\Psi^{-}_{I_2}$ suggests removing \textit{bumps} from $I_2$ to alter the classification category from \texttt{Toad} to \texttt{Frog}. Changing the classification result of $I_3$ from \texttt{Goat} to \texttt{Sheep} requires adding \textit{wool} and removing \textit{beard} and \textit{horns} from $I_3$, i.e., it needs both positive and negative fault-lines.}~\label{fig:intro2}
%\end{figure*}

\begin{figure*}[t]
\centering
  \includegraphics[width=\linewidth]{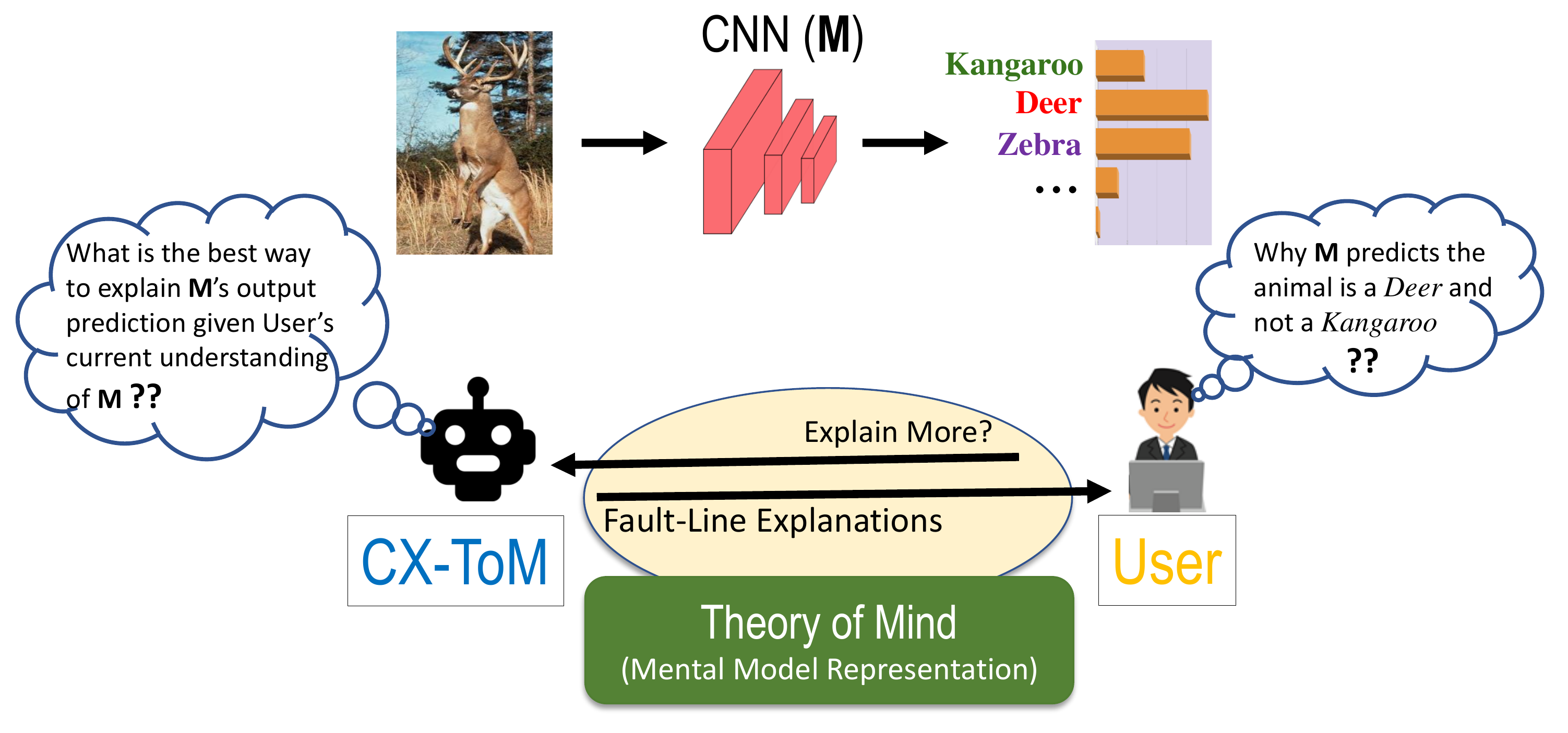}
  \caption{Example of a ToM based Fault-Line Selection Process: The interaction is conducted through a dialog where the user seeks explanations about CNN output predictions. CX-ToM picks an optimal fault-line as explanation based on user's (estimated) current understanding of model.}~\label{fig:intro2}
\end{figure*}

In addition, we propose novel counterfactual explanations called \textit{fault-lines} and show that they are superior to attention based explanations. Fault-lines are the high-level semantic aspects of reality that humans zoom in on when they imagine an alternative to it. More concretely, given an input image $I$ for which a CNN model $M$ predicts class $c_{pred}$, our fault-line based explanation identifies a \textit{minimal} set of semantic features, referred to as \textit{explainable concepts} (xconcepts), that need to be added to or deleted from $I$ in order to alter the classification category of $I$ by $M$ to another specified class $c_{alt}$~\citep{byrne2017counterfactual,kahneman1981simulation,akula2020cocox,8012316,byrne2002mental,byrne2007rational}. For example, let us consider a training dataset for an image classification task shown in Figure~\ref{fig:intro2} containing the classes \texttt{Dog}, \texttt{Thylacine}, \texttt{Frog}, \texttt{Toad}, \texttt{Goat} and \texttt{Sheep}, and a CNN based classification model $M$ which is trained on this dataset. In order to alter the model's prediction of input image $I_1$ from \texttt{Dog} to \texttt{Thylacine}, the fault-line ($\Psi^{+}_{I_1,c_{pred},c_{alt}}$) suggests adding \textit{stripes} to the \texttt{Dog}. We call this a positive fault-line (PFT) as it involves adding a new xconcept, i.e., \textit{stripedness}, to the input image.  Similarly, to change the model prediction of $I_2$ from \texttt{Toad} to \texttt{Frog}, the fault-line ($\Psi^{-}_{I_2,c_{pred},c_{alt}}$) suggests removing \textit{bumps} from the \texttt{Toad}. We call this a negative fault-line (NFT) as it involves subtracting xconcept, i.e., \textit{bumpedness}, from the input image. 

In most cases, both PFT and NFT are needed to successfully alter the model prediction. For example, in Figure~\ref{fig:intro2}, in order to change the model prediction of $I_3$ from \texttt{Goat} to \texttt{Sheep}, we need to add an xconcept \textit{wool} (PFT) to $I_3$ and also remove xconcepts \textit{beard} and \textit{horns} (NFT) from $I_3$. As we can see, these fault-lines can be directly used to make the internal decision making criteria of deep neural network transparent to both expert and non-expert users. For instance, we answer the question \textit{``Why does the model classify the image $I_3$ as \texttt{Goat} instead of \texttt{Sheep}?"} by using PFT $\Psi^{+}_{I_3,c_{pred},c_{alt}}$ and NFT $\Psi^{-}_{I_3,c_{pred},c_{alt}}$ as follows: ``Model thinks the input image is \texttt{Goat} and not \texttt{Sheep} mainly because \texttt{Sheep}'s feature \textit{woolly} is absent in $I_3$ and \texttt{Goat}'s features \texttt{beard} and \texttt{horns} are present in $I_3$". It may be noted that there could be several other features of \texttt{Sheep} and \texttt{Goat} that might have influenced the model's prediction. However, fault-lines only capture the most critical (minimal) features that highly influenced the model's prediction.

Note that fault-lines are \textbf{counter-factual} in nature, i.e., they provide a \textit{minimal} amount of information capable of altering a decision. This makes them easily digestible and practically useful for understanding the reasons for a model's decision~\citep{wachter2017counterfactual}. For example, consider the fault-line explanation for image $I_3$ in Figure~\ref{fig:intro2}. The explanation provides only the most critical changes (i.e., adding wool and removing beard and horns) required to alter the model's prediction from \texttt{Goat} to \texttt{Sheep}, though several other changes may be necessary. While there are recent works on generating pixel-level counter-factual and contrastive explanations~\citep{hendricks2018generating,dhurandhar2018explanations,Goyal2019counterfactual}, to the best of our knowledge, this is the first work to propose a method for generating explanations that are iterative, counter-factual as well as conceptual. 

% We identify two main challenges in generating a fault-line explanation, namely: (a) How to identify the set of xconcepts; and (b) How to select the most critical xconcepts that alter the model prediction from $c_{pred}$ to $c_{alt}$. In this work, we first propose a novel method to mine all the plausible xconcepts from the given dataset automatically.
% We then identify class-specific xconcepts by using
% %Given a class, some of these Xconcepts are more influential than others. 
% directional derivatives~\citepp{kim2018interpretability}.
% %to rank these Xconcepts by quantifying the degree to which they are useful in explaining a class. 
% Next, we pose the derivation of a fault-line as an optimization problem which selects a minimal set of these xconcepts to alter the model's prediction. 

It may be noted that there exists multiple fault-lines that could be used to explain model's decisions. In this work, we pick the most optimal fault-line, i.e. the one that is most influential and suitable given the user's current understanding of CNN model, by using Theory-of-Mind (ToM)~\citep{yoshida2008game,rabinowitz2018machine,pearce2014social,raileanu2018modeling,ramirez2011goal,edmonds2019tale,zhang2018visual}. 

\subsection{Example of a ToM based Fault-Line Selection Process}
Given an input image and two output categories, fault-lines show the most important features or attributes that influence model's decision in classifying the image as one among the two output categories. In most cases, there exists several thousands of output categories and it is impossible for the human user to verify the model's reasoning and behaviour by constructing a fault-line between all the possible pairs of output categories. Therefore, it is important for the model to automatically select the most important pair for constructing fault-line explanation that helps human user to quickly understand the model's strengths or weaknesses. CX-ToM addresses this by incorporating Theory-of-Mind framework which helps in explicitly tracking human user's beliefs. More concretely, at each turn in the dialog, we estimate the human's understanding of the CNN model and generate a most suitable fault-line explanation aimed at increasing human understanding (and therefore trust) of the model. It may be noted that we are not trying to estimate or build rich and dynamic true state of a human mind using ToM - a grand challenge for AI. Instead, similar to prior works on ToM~\citep{yoshida2008game,rabinowitz2018machine,pearce2014social,raileanu2018modeling,ramirez2011goal,zhu2020dark}, we cast ToM framework as a simple learning problem that enable us to better understand user preferences that improve the utility of the explanations. 

\begin{figure*}[t]
\centering
  \includegraphics[width=\linewidth]{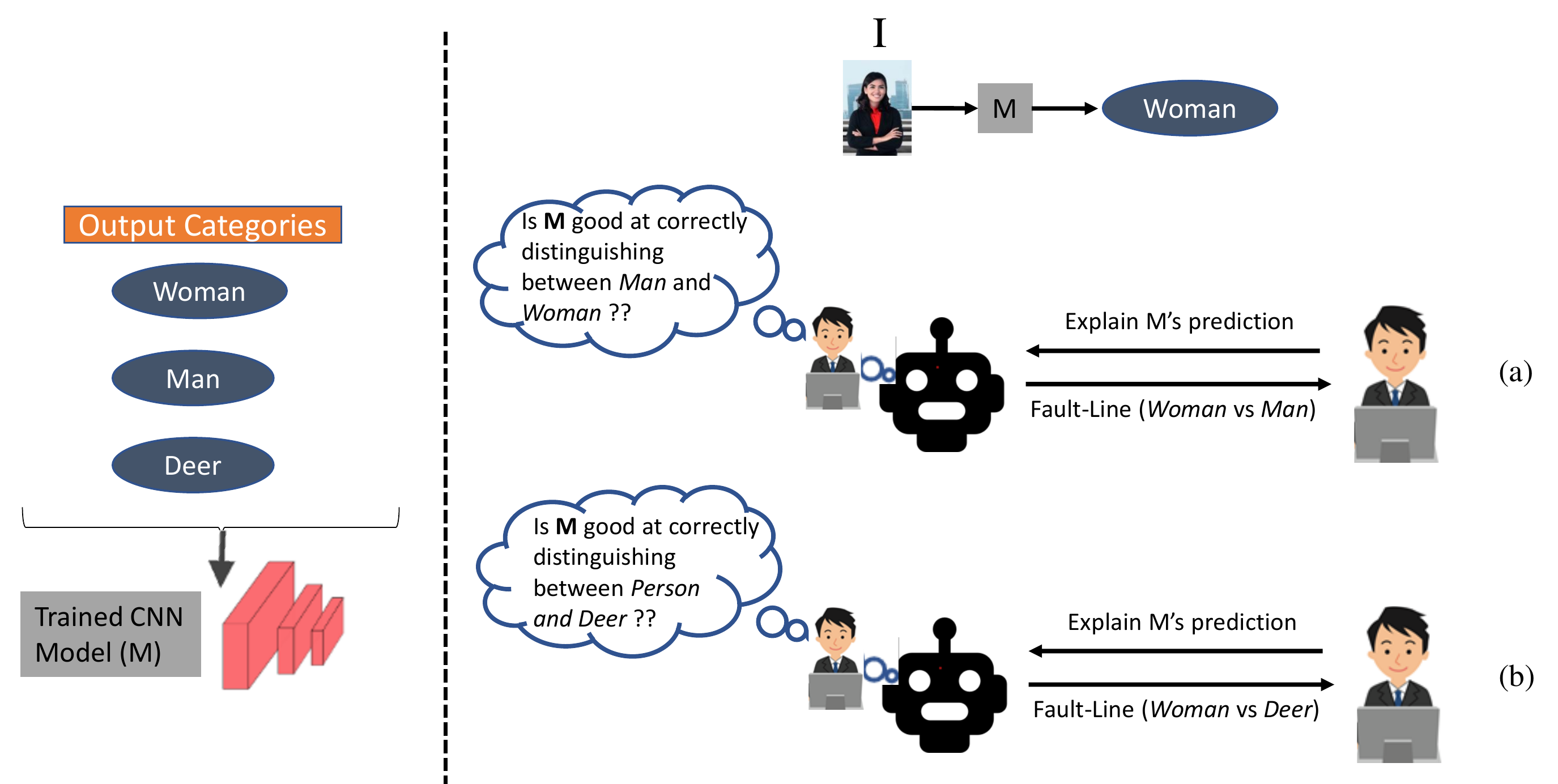}
  \caption{We select a fault-line explanation by estimating human user's current understanding of the model. For example, consider the first scenario (a), where CX-ToM estimates that user is not confident in model's ability to correctly classify between \texttt{Woman} and \texttt{Man}. Therefore, CX-ToM generates a fault-line explanation using the output categories \texttt{Woman} and \texttt{Man}. Whereas in the second scenario (b), CX-ToM thinks that user do not trust model's ability in correctly classifying \texttt{Person} and therefore shows a fault-line explanation using categories \texttt{Woman} and \texttt{Deer}.}~\label{fig:intro3}
\end{figure*}

For example, consider an input image shown in Figure~\ref{fig:intro3}, where the CNN model classifies the image as a \texttt{Woman}. The possible output categories are \texttt{Woman}, \texttt{Man}, and \texttt{Deer}. Generating a most suitable fault-line explanation to help user understand the model's reasoning process requires understanding the human user's current understanding of the model. If user knew that the model performs well at identifying \texttt{Person} but not very certain in its ability to correctly classify between \texttt{Man} and \texttt{Woman}, then the fault-line for the class pair $<$\texttt{Woman}, \texttt{Man}$>$ is a most appropriate explanation for the user. However, if the user is not certain about model's ability in correctly classifying \texttt{Person}, then the fault-line for the class pair $<$\texttt{Woman}, \texttt{Deer}$>$ is the most appropriate explanation. 

In summary, CX-ToM constructs explanations in the dialog using fault-lines and picks an optimal explanation based on ToM. We perform extensive human study experiments to demonstrate the effectiveness of our approach in improving human understanding of the underlying classification model. Through our ablations and human studies, we show that our CX-ToM explanations significantly outperform the baselines (i.e., attribution techniques and pixel-level counterfactual explanations) in terms of qualitative and quantitative metrics such as Trust and Explanation Satisfaction~\citep{hoffman2018metrics}.

\subsection{How is human trust measured in CX-ToM?}\label{sec:1.2}
In this work, we focus mainly on measuring and increasing \textbf{Justified Positive Trust} (JPT) and \textbf{Justified Negative Trust} (JNT)~\citep{hoffman2018metrics} in image classification models. We measure JPT and JNT by evaluating the human’s understanding
 of the machine’s (M) decision-making process. For example, if the image classification model $M$ predicts images in the set $C$ correctly and makes incorrect decisions on the images in the set $W$. Intuitively, JPT will be computed as the percentage of images in $C$ that the human subject felt $M$ would correctly predict. Similarly, JNT (also called as mistrust), will be computed as the percentage of images in $W$ that the human subject felt $M$ would fail to predict correctly. In other words, given an image, justified trust evaluates whether the users could reliably predict the model’s output decision. Note that this definition of justified trust is domain generic and can be easily adapted to any task. For example, in an AI-driven clinical world, our definitions of JPT and JNT can effectively measure how much doctors and patients understand the AI systems that assist in clinical decisions. 

% We perform extensive human study experiments to demonstrate the effectiveness of our approach in improving human understanding of the underlying classification model. Through our human studies, we show that our CX-ToM significantly outperform the baselines (i.e., attribution techniques and pixel-level counterfactual explanations) in terms of qualitative and quantitative metrics such as Justified Trust and Explanation Satisfaction~\citep{hoffman2018metrics}.

% seeks to automatically identify human-friendly xconcepts. However, they use segmentation methods to identify xconcepts, whereas we use Grad-CAM~\citep{selvaraju2017grad} based localization maps. Moreover, their explanations are not counter-factual unlike our fault-line based explanations. 
%Concurrent to our work, recent work by \citep{ghorbani2019automating} also seeks to automatically identify human-friendly xconcepts. However, they use segmentation methods to identify xconcepts, whereas we use Grad-CAM~\citep{selvaraju2017grad} based localization maps. Moreover, their explanations are not counter-factual unlike our fault-line based explanations. 

Our contributions are summarized below: 
\begin{itemize}
 \item We introduce a new XAI framework based on Theory-of-Mind and counterfactual explanations.
 \item We present a ToM based approach to automatically select the most important pair of output categories for constructing fault-line explanation.
 \item We show that CX-ToM XAI framework qualitatively and quantitatively outperform baselines in improving human understanding of the classification model.
\end{itemize}

The remainder of this paper is organized as follows. Section 2 reviews the previous work done in explaining image classification models. Section 3 introduces our CX-ToM explanation framework. In section 4, we present our experimental results and section 5 draws conclusions and points to future directions for research. 

% \section{Background}
% \subsection{Why XAI is an important problem to solve?}
% XAI models, through explanations, make the underlying inference mechanism of AI systems transparent and interpretable to expert users (system developers) and non-expert users (end-users)~\citep{lipton2016mythos,ribeiro2016should,miller2017explanation,hoffman17explanation,lipton1990contrastive,vstrumbelj2011general}. Explanations play a key role in integrating AI machines into our daily lives, i.e. XAI is essential to increase social acceptance of AI machines (see Figure~\ref{fig:fig1}). As the decision making is being shifted from humans to machines, \textbf{transparency} and \textbf{interpretability} achieved with reliable explanations is central to solving AI problems such as the following: 
% \begin{enumerate}
% \item Safety~\citep{mandel2019let} (e.g. \textit{How to operate self-driving cars safely?})
% \item Bias \& Fairness~\citep{bellamy2018ai} (e.g. \textit{How to detect and mitigate bias in ML models?})
% \item Justified Human Trust in ML models~\citep{siau2018building} (e.g. \textit{How to trust the output of AI systems to inform our decisions?})
% \item Model Debugging~\citep{hall2019guidelines} (e.g. \textit{How to improve my model by identifying points of model failure?})
% \item Ethics~\citep{vakkuri2019ethically} (e.g. \textit{How to ensure that ML models reflect our values?})
% \end{enumerate}

\section{Related Work}
The importance of generating explanations or justifications of decisions made by an AI system has been emphasized and widely explored in numerous works over the past decades~\citep{alang2017turns,bornstein2016artificial,champlin2017ai,bach2015pixel,shrikumar2017learning,zhou2016learning,berry1987explanation,biran2017explanation,darlington2013aspects,doshi2017roadmap,doshi2017towards,goodman2017european,hoffman2017taxonomy,hoffman2017explaining,keil2006explanation,kulesza2010explanatory,kulesza2011my,moore1990pointing,walton2004new,walton2007dialogical,walton2011dialogue,sheh2017did,sheh2018defining,tapaswi2016movieqa,williams2016axis,agarwal2018structured,akula2019natural,akula2019x,akula2019visual,akula2019explainable,gupta2016desire,akula2021measuring,akula2018analyzing,bivens2017cognitive,zhang2020mining,zhang2020extracting,zhang2019interpretable}. Most prior work in explaining CNN's predictions has focused on generating explanations using feature visualization and attribution.\\ 

\noindent
\textbf{Feature visualization} techniques typically identify  qualitative interpretations of  features used for making predictions or decisions. 
For example, gradient ascent optimization is used in the image space to visualize the hidden feature layers of unsupervised deep architectures~\citep{erhan2009visualizing}. Also, convolutional layers are visualized by reconstructing the input of each layer from its output~\citep{zeiler2014visualizing}. Recent visual explanation models seek to jointly classify the image and explain why the predicted class label is appropriate for the image~\citep{hendricks2016generating}. Other related work includes a visualization-based explanation framework for Naive Bayes classifiers~\citep{szafron2003explaining}, an interpretable  character-level language models for analyzing the predictions in RNNs~\citep{karpathy2015visualizing}, and an interactive visualization for facilitating analysis of RNN hidden states \citep{strobelt2016visual}.

\noindent
\textbf{Attribution} is a set of techniques that highlight pixels of the input image (saliency maps) that most caused the output classification. Gradient-based visualization methods~\citep{zhou2016learning,selvaraju2017grad} have been proposed to extract image regions responsible for the network output. The LIME method proposed by~\citep{ribeiro2016should} explains  predictions of any classifier by approximating it locally with an interpretable model. SHAP~\citep{NIPS2017_7062}, another common attribution technique, uses shapley values to explain output predictions of a model for given input by computing the contribution of each feature to the prediction.  

There are few recent works in the XAI literature that go beyond the pixel-level explanations. For example, the TCAV technique proposed by \citep{kim2018interpretability} aims to generate explanations based on high-level user defined concepts. Contrastive explanations are proposed by \citep{dhurandhar2018explanations} to identify minimal and sufficient features to justify the classification result. \citep{Goyal2019counterfactual} proposed counterfactual visual explanations that identify how the input could change such that the underlying vision system would make a different decision.
More recently, few methods have been developed for building models which are intrinsically interpretable~\citep{zhang2018interpretable}. In addition, there are several works~\citep{miller2018explanation,hilton1990conversational,lombrozo2006structure} on the goodness measures of explanations which aim to assess the underlying characteristics of explanations. 

We further categorize above works on feature visualization and attribution as follows:
\subsection{\textbf{Intrinsic vs Post-hoc Explanations}}
Explanations that are derived (or understood) directly from the model's internal representation or the output parse structure are called as Intrinsic Explanations~\citep{doshi2017towards,zhang2018interpretable,zhang2017interpretable,stone2017teaching}. For example, the reasoning behind the predictions made by linear regression models, decision trees, and And-Or Graphs~\citep{li2013modeling,zhang2017mining} is easier to understand without using any external XAI models and hence are considered as intrinsically explainable. These models, due to their simple structure, typically do not fare well in terms of performance compared to black-box models such as deep neural nets. Majority of the work in XAI is focused on generating post-hoc~\citep{lei2016rationalizing,ribeiro2016should,kim2018interpretability,kim2014bayesian,wang2016bayesian,kim2015mind} explanations where an external XAI model is employed to explain the model. More recently, there are efforts in making the complex deep neural networks intrinsically explainable~\citep{zhang2017interpretable,zhang2019interpreting,zhang2018network}. For example, \citep{zhang2019interpreting} proposed a decision tree to encode decision modes in fully-connected layers and thereby quantitatively explain the logic for each CNN prediction.

\subsection{\textbf{Model-agnostic vs Model-specific Explanations}}
Explainable AI models that do not require CNN model specific details (for example, weights of CNN) for generating explanations are called as model-agnostic models~\citep{ribeiro2018anchors}. In other words, they simply analyze the dependencies of input features against the output predictions to explain the model's decision. It may be noted that intrisinc explanations are typically model-specific whereas post-hoc XAI models are model-agnostic. Several XAI works belong to this category, to name a few:
\begin{enumerate}
\item \textit{Local Intepretable Model-Agnostic Explanation (LIME)} \citep{ribeiro2016should}. LIME produces attention map as explanation, generated through super-pixel based perturbation. Though LIME is a post-hoc model-agnostic model, it generates explanations by approximating the model (locally) with an intrinsic model-specific XAI model.
\item \textit{Contrastive Explanation Methods (CEM)} \citep{dhurandhar2018explanations}. CEM provides contrastive explanations by identifying pertinent positives and pertinent negatives in the input image.
\item \textit{Counterfactual Visual Explanations (CVE)} \citep{Goyal2019counterfactual}.  CVE provides counterfactual explanation describing what changes to the situation would have resulted in arriving at the alternative decision. 
\end{enumerate}

\subsection{\textbf{Human Interpretable Explanations (Concept Activation Vectors)}}
Most XAI models represent the explanations using attention maps (saliency). However, these explanations are difficult for humans to understand. For example, authors in \citep{DBLP:journals/corr/abs-1902-10186} considered NLP tasks (text classification, natural language inference (NLI), and question answering) to show that attention mechanism is not useful for humans. Therefore, there is a dire need to represent and generate human-friendly explanations. Recent work by \citep{kim2018interpretability} presents a first step towards this goal. They propose a technique called TCAV that takes the the user defined concept ($X$) represented using a set of example images and maps it to the activation space of any given layer $l$ in the network. It then constructs a vector representation of each concept, called CAV (denoted as $v_X$), by using a direction normal to a linear classifier trained to distinguish between the concept activations from the random activations. The sensitivity of network predictions towards a concept is gauged by computing directional derivatives ($S_{c,X}$) to produce estimates of how important the concept $X$ was for a CNN's prediction of a target class $c$, e.g. how important is the concept \texttt{stripedness} for predicting the zebra class.
\begin{equation}
    \begin{aligned}
        S_{c,X} = \nabla g_c (f(I)) \cdot v_X
    \end{aligned}
\end{equation}
where $g_c$ denote classifier component of CNN that takes output of $f$ and predicts log-probability of output class $c$. Because TCAV provides explanations using high-level concepts, it is expected to achieve higher human trust and reliance values compared to the attention based explanations~\citep{selvaraju2017grad,ribeiro2016should}.

\subsection{\textbf{Proxy or Surrogate Models}}
A Proxy or surrogate model is a simpler interpretable model that approximates the behaviour of the complex model~\citep{ribeiro2016should,alvarez2018robustness,sato2001rule,augasta2012reverse}. It reduces the complexity of the original model but produces similar output estimates. Most surrogate XAI models are model-agnostic. A surrogate model that is trained to explain individual instances is referred to as local surrogate model.  For example, LIME~\citep{ribeiro2016should} approximates model with a local linear model that serves as a surrogate for the model in the neighborhood of the input. Similarly, authors in ~\citep{sato2001rule,zhang2017mining} locally approximate neural networks with decision trees. This notion of using proxy models is also referred to as Knowledge Distillation~\citep{hinton2015distilling,hernandez2018deep,polino2018model} and Rule Extraction~\citep{zilke2016deepred}.

\subsection{\textbf{Perturbation Analysis}}
Perturbation analysis helps in measuring the feature importance for the predictions made by model~\citep{fisher2018model,moosavi2017universal}. The assumption here is that model's confidence in the prediction will be low if an important feature has been removed (or masked) after perturbing the input features. Adversarial analysis~\citep{goodfellow2014explaining} and Probing techniques~\citep{clark2019does} are few popular techniques for perturbation analysis.

\subsection{\textbf{Counterfactual Explanations}}
Counterfactual (and Contrastive) explanations provide a \textit{minimal} amount of information capable of altering a model's decision. In other words, they aim at describing the causal situations such as ``What would be the output of model if X had not occurred?". This makes them easily digestible and practically useful for understanding the reasons for a model's decision~\citep{pedreschi2018open,wachter2017counterfactual,Goyal2019counterfactual,van2019interpretable}.

For example, \citep{fong2017interpretable} propose a counterfactual reasoning framework to find the part of an image most responsible for a classifier decision.  This saliency based explanation framework helps in understanding where the model looks by discovering which parts of an image most affect its output score when perturbed. \citep{Goyal2019counterfactual} proposes a counterfactual explanation framework to identify how the input image could be changed such that the model would output a different specified class. To do this, they select a distractor image that the model predicts as class $c_1$ and identify spatial regions such that replacing the identified region in input image with the regions from distractor image would push the model towards classifying I as $c_2$. Contrastive explanations are proposed by \citep{dhurandhar2018explanations} to identify minimal and sufficient features to justify the classification result. Unlike these prior counterfactual explanation frameworks which mainly focus on pixel-level explanations (viz. saliency maps), our proposed ToM based counterfactual explanations, i.e. fault-lines, are \textbf{concept-level} explanations. Pixel-level explanations are not effective at human scale, whereas concept level explanations are effective, less ambiguous, and more natural for both expert and non-expert users in building a mental model of a vision system~\citep{kim2018interpretability}.
Moreover, with conceptual explanations, humans can easily generalize their understanding to new unseen instances/tasks.
 
\subsection{\textbf{Partial Dependence Plots}}
Partial dependence plots (PD) is a model-agnostic XAI technique that helps in understanding the relationships between one or more input variables as well as marginal effect of a given variable on a model's decision~\citep{friedman2001greedy,hastie01statisticallearning,molnar2019interpretable}.

\subsection{\textit{Class Activation Mapping (CAM)}} CAM produces attention map as explanation, i.e. it highlights the important regions in the image for predicting a target output. Gradient-weighted Class Activation Mapping (Grad-CAM) \citep{selvaraju2017grad} uses the gradients of target class flowing into the final convolutional layer to produce attention map as explanation. Layer-wise Relevance Propagation (LRP) \citep{bach2015pixel} generates attention map by propagating classification probability backward through the network and then calculates relevance scores for all pixels. SmoothGrad \citep{smilkov2017smoothgrad} produces attention map as explanation by adding gaussian noise to the original image and then calculating gradients multiple times and averaging the results.

\section{CX-ToM Framework}

In this section, we first demonstrate the importance of ToM based explanations by designing a collaborative task-solving game for visual recognition (Sec 3.1). We next present the fault-lines as an alternative to attention based explanations (Sec 3.2). Finally, we detail our CX-ToM model which integrates both ToM and fault-lines into one single explanation framework (Sec 3.3).

\begin{figure*}[t]
\centering
  \includegraphics[width=0.95\linewidth,height=0.45\linewidth]{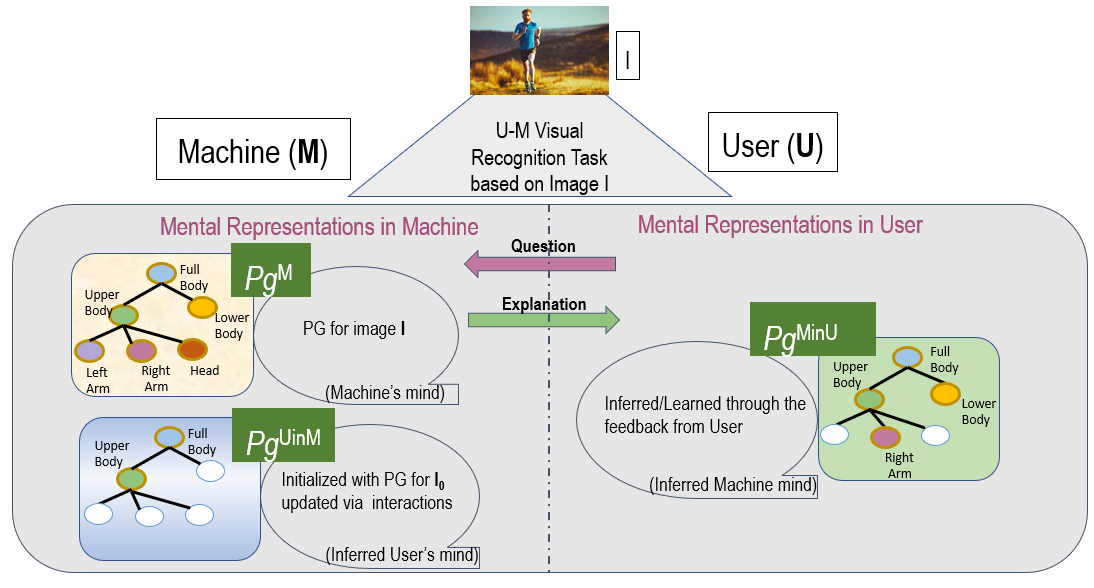}
  \caption{Our collaborative ToM based interaction framework for a visual recognition task consists of three distinct parse graphs ($pg$'s): $pg^M$ representing the machine's interpretation of the image, $pg^{UinM}$ --- the human's mind as inferred by the machine; and $pg^{MinU}$ --- the machine's mind as inferred by the human. Nodes of a parse graph represent objects and parts appearing in the image, and edges represent spatial relationships of the objects. We use ToM to optimize explanations so as to reduce a difference among the three parse graphs.}~\label{fig:xtom_idea}
\end{figure*}

\begin{figure*}[t]
\centering
  \includegraphics[width=0.76\linewidth,height=0.4\linewidth]{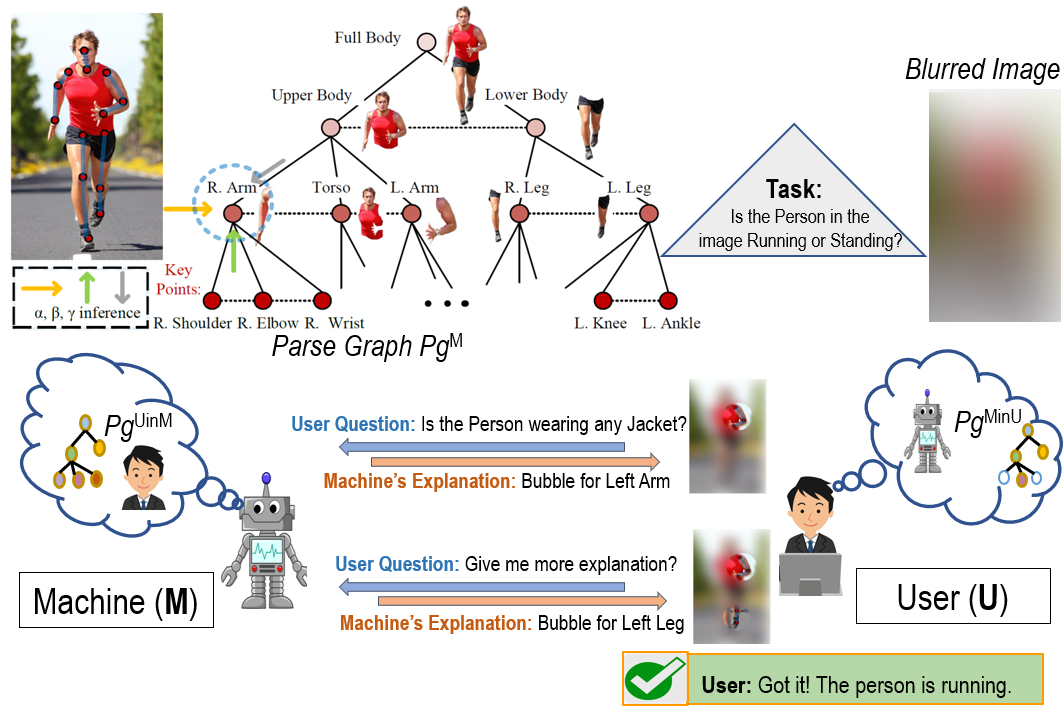}
  \caption{An example of the first phase of our ToM based collaborative game aimed at estimating $pg^{UinM}$: The user is shown a blurred image and given a task to recognize if the person in the image is running or walking. The machine has access to the original (unblurred) image and $pg^M$. The user then asks questions regarding objects and parts in the image. Using the detections in $pg^M$, machine provides visual explanations as ``bubbles'' that reveal the corresponding image parts in the blurred image. The generated explanations are used to update $pg^{UinM}$.}~\label{fig:xtom}
\end{figure*}

\subsection{Importance of ToM}
We test the importance of ToM for providing effective explanations by designing a collaborative task-solving game for visual recognition. In this game, the machine is given an original image and is supposed to detect and localize objects and parts of interest or a human activity appearing in the image. The user is given a blurred version of the original image, and the user seeks the machine's help essentially through the explanations generated by the machine in order to recognize objects/parts in the blurred image. This provides a unique collaborative setting where the system is motivated to provide human-understandable explanation for its visual recognition and the user is motivated to seek the system's recognition and explanation to help his/her own understanding. To facilitate this collaborative interaction, we use ToM to explicitly model mental states of visual understanding (``minds'') of the machine and user using parse graphs ($pg$) in the form of And-Or Graph (AOG)~\citep{zhu2007stochastic}. In a $pg$, nodes represent objects and parts detected in the image, and edges represent spatial relationships identified between the objects. As shown in Figure~\ref{fig:xtom_idea}, we have three main components as part of this interaction: 
\begin{itemize}
    \item A {\bf Performer} that generates image interpretations (i.e., machine's mind represented as $pg^M$) using a set of computer vision algorithms;
    \noindent
    \item An {\bf Explainer} that generates maximum utility explanations in a dialog with the user by accounting for $pg^{M}$ and  $pg^{UinM}$ using reinforcement learning;
    \noindent
    \item An {\bf Evaluator} that quantitatively evaluates the effect of explanations on the human's understanding of the machine's behaviors (i.e., $pg^{MinU}$) and measures human trust by comparing $pg^{MinU}$ and $pg^{M}$. 
\end{itemize}

The game consists of two phases. In the first phase, %where the user needs to understand a blurred image based on machine's recognition and explanation.  %for the machine to explain to the user (i.e., \textbf{Explainer}), 
the user is shown a blurred image and given a task to recognize what the image shows. The machine has access to the original (unblurred) image and the machine's (i.e. \textbf{Performer's}) inference result $pg^M$. The user is allowed to ask questions regarding objects and parts in the image that the user finds relevant for his/her own recognition task. Using the detected objects and parts in $pg^M$, \textbf{Explainer} provides visual explanations to the user, as shown in Figure~\ref{fig:xtom}. This process allows the machine to infer what the user sees and iteratively update $pg^{UinM}$, and thus select an optimal explanation at every turn of the game. Optimal explanations generated by the \textbf{Explainer} are the key to maximize the human trust in the machine. 
%Through these explanations, we hope the user will develop an increasing understanding of the performer's behaviors.   
%While the first phase is for a user and a machine to engage in a collaborative task, 

The second phase is specifically designed for evaluating whether the explanation provided in the first phase helps the user understand the system behaviors. 
%for the user to evaluate the machine's explanation ability based on his/her experience in the first phase. More specifically, in the second phase, 
The \textbf{Evaluator} shows a set of original (unblurred) images to the user that are similar to (but different from) the ones used in the first phase of the game (i.e., the set of images shows the same class of objects or human activity). The user is then given a task to predict in each image the locations of objects and parts that would be detected by the machine (i.e., in $pg^M$) according to his/her understanding of the machine's behaviors. Based on the human predictions, the \textbf{Evaluator} estimates $pg^{MinU}$ and quantifies human trust in the machine by comparing $pg^{MinU}$ and $pg^{M}$.
% aids user in solving collaborative tasks by providing optimal explanations. User is expected to improve his understandability and predictability of machine's (i.e. \textbf{Performer's}) internal computational processes by looking at the machine's explanations in this phase. In the second phase, \textbf{Evaluator} evaluates the extent to which user is able to improve his understandability and predictability of the model's underlying inference processes.
% Specifically, in the first phase of an X-ToM game, 

The three minds $pg^{M}$, $pg^{MinU}$, and $pg^{UinM}$ are sub-graphs of an And-Or Graph (AOG) defining all objects, parts, and their relationships and attributes of the visual domain considered. The AOG uses AND nodes to represent decompositions of human body parts into subparts, and OR nodes for alternative decompositions. 
Each node is characterized by attributes that pertain to the corresponding human body part, including the pose and action of the entire body. Also, edges in the AOG capture hierarchical and contextual relationships of the human body parts. Our AOG-based performer uses three inference processes $\alpha$, $\beta$ and $\gamma$ at each node. Figure~\ref{fig:xtom} shows an example part of the AOG relevant for human body pose estimation~\citep{park2016attribute}. The $\alpha$ process detects nodes (i.e., human body parts) of the AOG directly based on image features, without taking advantage of the surrounding context. The $\beta$ process infers nodes of the AOG by binding the previously detected children nodes in a bottom-up fashion, where the children nodes have been detected by the $\alpha$ process (e.g., detecting human's upper body from the detected right arm, torso, and left arm).  Note that the $\beta$ process is robust to partial object occlusions as it can infer an object from its detected parts. 
The $\gamma$ process infers a node of the AOG top-down from its previously detected parent nodes, where the parents have been detected by the $\alpha$ process (e.g., detecting human's right leg from the detected outline of the lower body). The parent node passes contextual information so that the performer can detect the presence of an object or part from its surround. Note that the $\gamma$ process is robust to variations in scale at which objects appear in images.

\begin{figure*}[t]
\centering
  \includegraphics[width=0.8\linewidth,height=0.5\linewidth]{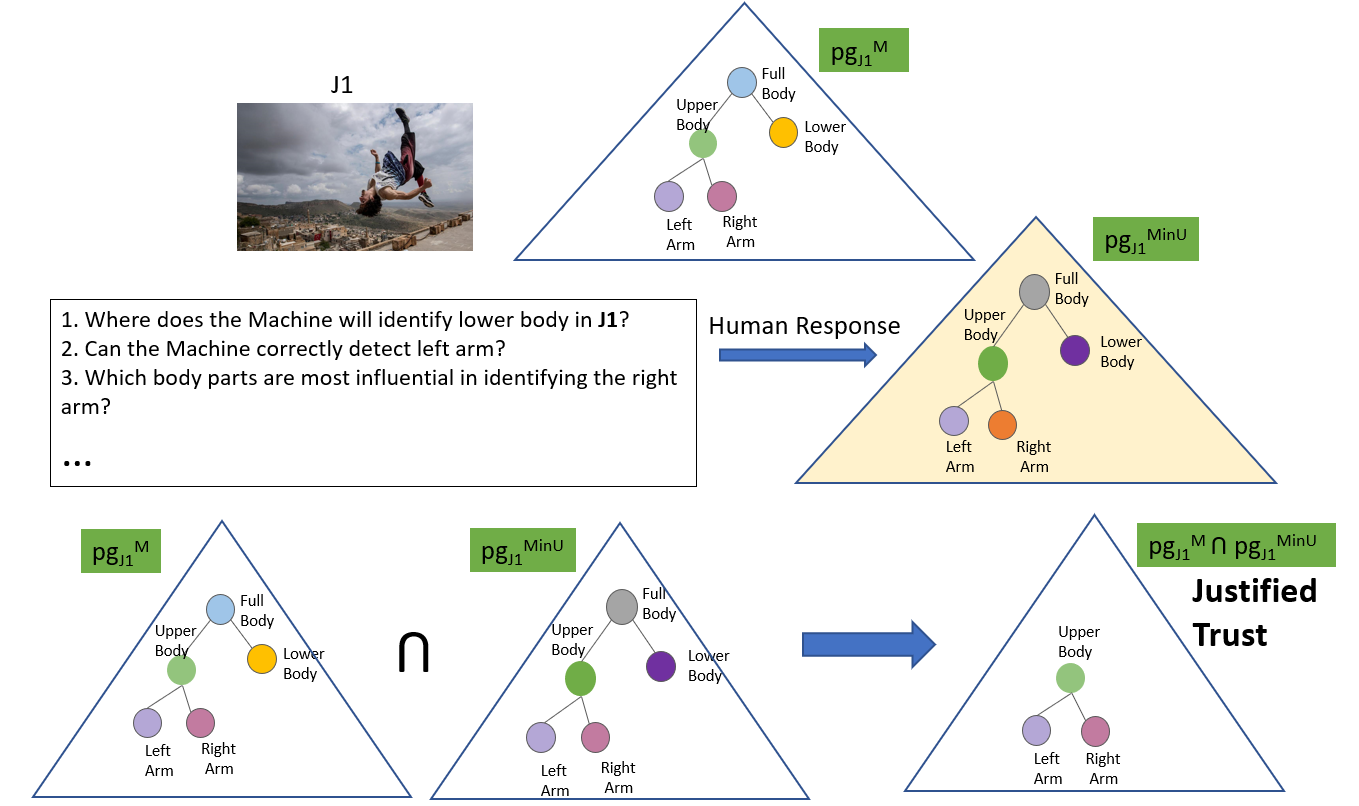}
  \caption{An example of second phase of ToM game where we estimate $pg^{MinU}$ and also quantitatively compute justified trust.}~\label{fig:xtom_trust}
\end{figure*}

The explainer, in the first phase of the game, makes the underlying $\alpha$, $\beta$, and $\gamma$ inference process of the performer more transparent to the human through a collaborative dialog. At one end, the explainer is provided access to an image and the performer's inference result $pg^{M}$ on that image. At the other end, the human is presented a blurred version of the same image, and asked to recognize a body part, or pose, or human action depicted (e.g., whether the person is running or walking). To solve the task, the human may ask the explainer various ``what'', ``where'' and ``how'' questions (e.g.,  ``Where is the left arm in the image''). We make the assumption that the human will always ask questions that are related to the task at hand so as to solve it efficiently. As visual explanations, we use ``bubbles" \citep{gosselin2001bubbles}, where each bubble reveals a circular part of the blurred image to the human. The bubbles coincide with relevant image parts for answering the question from the human, as inferred by the performer in $pg^{M}$. For example, a bubble may unblur the person's left leg in the blurred image, since that image part has been estimated in $pg^{M}$ as relevant for recognizing the human action ``running'' occurring in the image.\\
The second phase of the X-ToM game serves to assess the effect of the explainer on the human's understanding of the performer. This assessment is conducted by the evaluator. The human is presented with a set of (unblurred) images that are different from those used in the first phase. For every image, the evaluator asks the human to predict the performer's output. The evaluator poses multiple-choice questions and the user clicks on one or more answers. As shown in Figure~\ref{fig:xtom_trust}, we design these questions to capture different aspects of human's understanding of $\alpha$, $\beta$ and $\gamma$ inference processes in the performer. Based on responses from the human, the evaluator estimates $pg^{MinU}$. By comparing $pg^{MinU}$ with the actual machine's mind $pg^{M}$ (generated by the performer), we have defined the following metrics to quantitatively assess human trust~\citep{hoffman17explanation,hoffman2010metrics,hoffman2018metrics,miller2018explanation} in the performer:\\
 (1) \textit{Justified Positive and Negative Trust:} It is possible for humans to feel positive trust with respect to certain tasks, while feeling negative trust (i.e. mistrust) on some other tasks. The positive and negative trust can be a mixture of justified and unjustified trust~\citep{hoffman17explanation,hoffman2018metrics}. We compute justified positive trust (JPT) and negative trust (JNT) as follows:
 \begin{align}
\nonumber
\text{JPT} &= \dfrac{1}{N}\displaystyle\sum_{i}\displaystyle\sum_{z=\alpha,\beta,\gamma}\Delta \text{JPT}(i,z),\\
\nonumber
\Delta \text{JPT}(i,z)  &= \displaystyle\dfrac{\|pg^{MinU}_{i,z,+} \cap pg^{M}_{i,+}\|}{\|pg_{i,+}^{M}\|},\\
\nonumber
\text{JNT} &= \dfrac{1}{N}\displaystyle\sum_{i}\displaystyle\sum_{z=\alpha,\beta,\gamma}\Delta \text{JNT}(i,z),\\
\nonumber
\Delta \text{JNT}(i,z)  &= \displaystyle\dfrac{\|pg^{MinU}_{i,z,-} \cap pg^{M}_{i,-}\|}{\|pg_{i,-}^{M}\|},
\end{align}
\normalsize{}

where $N$ is the total number of games played. $z$ is the type of inference process. $\Delta \text{JPT}(i,z)$, $\Delta \text{JNT}(i,z)$ denote the justified positive and negative trust gained in the $i$-{th} turn of a game on the $z$ inference process respectively. $pg^{MinU}_{i,z,+}$ denotes nodes in $pg^{MinU}_{i}$ for which the user thinks the performer is able to accurately detect in the image using the $z$ inference process. Similarly, $pg^{MinU}_{i,z,-}$ denotes nodes in $pg^{MinU}_{i}$ for which the user thinks the performer would fail to detect in the image using the $z$ inference process. $\|pg\|$ is the size of $pg$. Symbol $\cap$ denote the graph intersection of all nodes and edges from two $pg$'s.\\
(2) \textit{Reliance:} Reliance (Rc) captures the extent to which a human can accurately predict the performer's inference results without over- or under-estimation. In other words, Reliance is proportional to the sum of JPT and JNT.
\begin{align}
\nonumber
\text{Rc} &= \dfrac{1}{N}\displaystyle\sum_{i}\displaystyle\sum_{z=\alpha,\beta,\gamma}\Delta \text{Rc}(i,z),\\
\nonumber
\Delta\text{Rc}(i,z) &= \displaystyle\dfrac{\|pg^{MinU}_{i,z} \cap pg^{M}_{i,z}\|}{\|pg_{i}^{M}\|}.
\end{align}
\normalsize{}
We deployed the ToM game on the Amazon Mechanical Turk (AMT) and trained the Explainer through the interactions with turkers. All the turkers have a bachelor’s degree or higher. We used three visual recognition tasks in our experiments, namely, human body parts identification, pose estimation, and action identification. We used 1000 images randomly selected from Extended Leeds Sports (LSP) dataset~\citep{johnson2010clustered}. Each image is used in all the three tasks. During training, each trial consists of one ToM game where a turker solves a given task on a given image. We restrict Turkers from solving a task on an image more than once. In total, about 2400 unique workers contributed in our experiments. We performed off-policy updates after every 200 trials, using Adam optimizer~\citep{kingma2014adam} with a learning rate of 0.001 and gradients were clipped at [-5.0, 5.0] to avoid explosion. We used $\epsilon$-greedy policy, which was annealed from 0.6 to 0.0. We stopped the training once the model converged. Using our Evaluator module, we conduct human subject experiments to assess the effectiveness of the ToM Explainer, that is trained on AMT, in increasing human trust through explanations. We recruited 120 human subjects from our institution's Psychology subject pool~\footnote{These experiments were reviewed and approved by our institution's IRB.}. We applied between-subject design and randomly assigned each subject into one of the three groups. One group used ToM Explainer, and two groups used the following two baselines respectively: 

\begin{itemize}
\item {\bf $\Omega_{\text{QA}}$}: we measure the gains in human trust only by revealing 
the answers for the tasks without providing any explanations to the human.

\item {\bf $\Omega_{\text{Salience}}$}: in addition to the answers, we also provide saliency maps generated using attribution techniques to the human as explanations~\citep{zhou2016learning,selvaraju2016grad}.
%\textcolor{blue}{jyc: in this setting, did you use RL to train the model to show Attention map? How was the Attnmap built upon the internal representation of pg? Just wondering whether the trust metrics are fairly defined for this setting. }

\end{itemize} 

Within each group, each subject will first go through an introduction phase where we introduce the tasks to the subjects. Next, they will go through familiarization phase where the subjects become familiar with the machine's underlying inference process (Performer), followed by a testing phase where we apply our trust metrics and assess their trust in the underlying Performer.

\begin{figure*}[t]
\centering
  \includegraphics[width=0.8\linewidth,height=7cm]{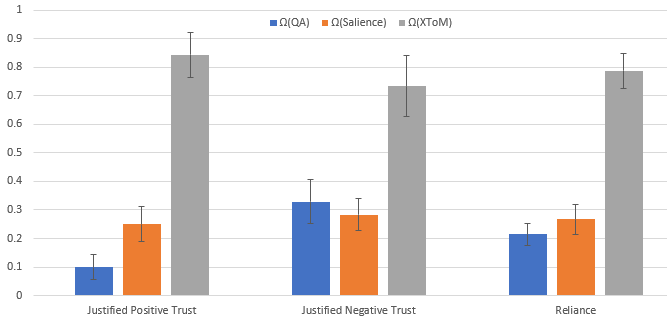}
  \caption{Gain in Justified Positive Trust, Justified Negative Trust and Reliance: our ToM framework (denoted as X-ToM) vs baselines (QA, Saliency Maps). Error bars denote standard errors of the means.}~\label{fig:trust_bars}
\end{figure*}

Figure~\ref{fig:trust_bars} compares the justified positive trust (JPT), justified negative trust (JPT), and Reliance (Rc) of ToM Explainer with the baselines.
%\textcolor{blue}{jyc: in each evaluation task, did the evaluator ask subjects questions for every node/edge in the machine's model? so the machine's mind and the machine in human's mind have the same size of pg?
%Shouldn't $\Delta \text{PT}(i,z) + \Delta \text{NT}(i,z) \leq 1$? }
As we can see, JPT, JNT and Rc values of ToM based framework are significantly higher than $\Omega_{\text{QA}}$ and $\Omega_{\text{Salience}}$ ($p < 0.01$). \textit{Also, it should be noted that attribution techniques ($\Omega_{\text{Salience}}$) did not perform any better than the $\Omega_{\text{QA}}$ baseline where no explanations are provided to the user}. This could be attributed to the fact that, though saliency maps help human subjects in localizing the region in the image based on which the performer made a decision, they do not necessarily reflect the underlying inference mechanism. In contrast, ToM Explainer makes the underlying inference processes ($\alpha$, $\beta$, $\gamma$) more explicit and transparent and also provides explanations tailored for individual user's perception and understanding. Therefore ToM explanations leads to the significantly higher values of JPT, JNT and Rc, confirming our hypothesis that ToM helps in providing effective explanations to the user.

\subsection{Fault-Lines as an alternative to Attention based Explanations}

\begin{figure*}[t]
\centering
  \includegraphics[width=0.9\linewidth]{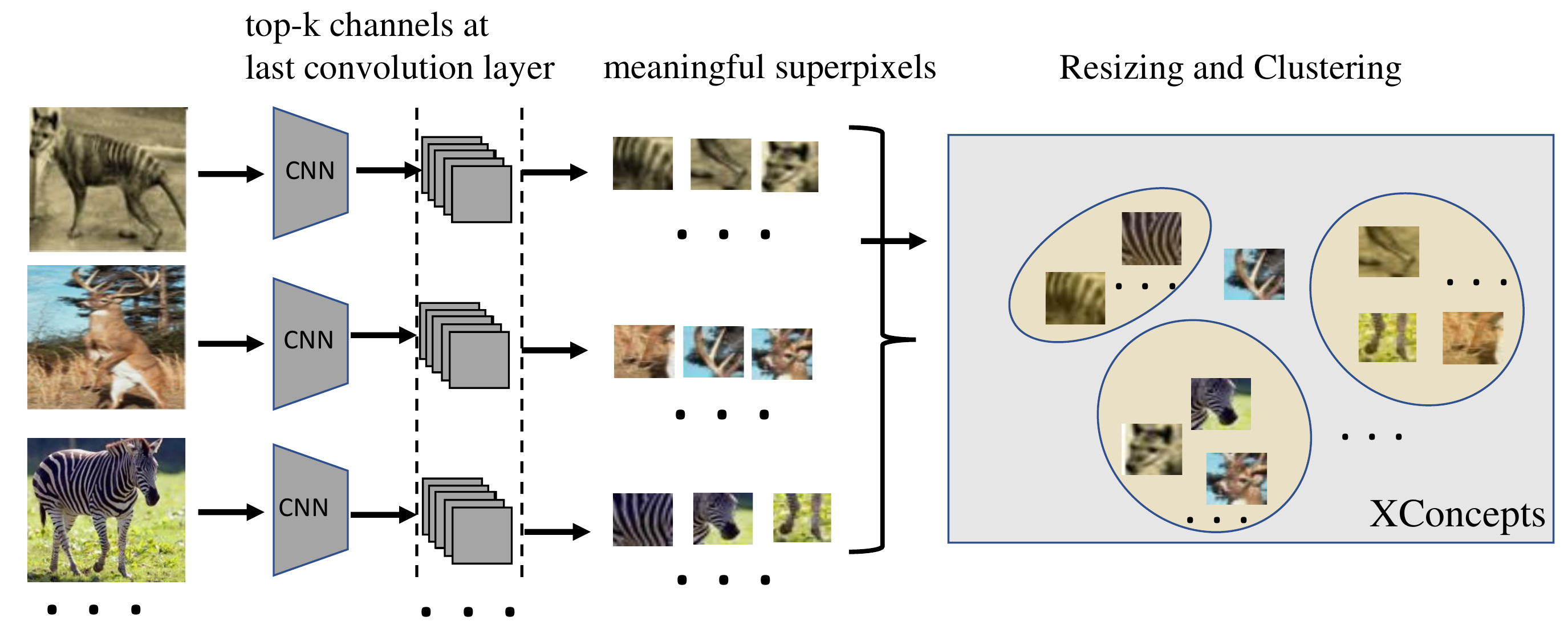}
  \caption{We consider feature maps from the last convolutional layer as instances of xconcepts and obtain their localization maps (i.e., superpixels) by computing the gradients of the output with respect to the feature maps. We select highly influential superpixels and then apply K-means clustering with outlier removal to group these superpixels into clusters where each cluster represents an xconcept.}~\label{fig:model1}
  %\vspace{-10pt}
\end{figure*}

In this section, we detail our ideas and methods for generating fault-line explanations. Without loss of generality, we consider a pre-trained CNN ($M$) for image classification. Given an input image $I$, the CNN predicts a log-probability output $\log P \left({\bf Y} \middle\vert I\right)$ over the output classes {\bf Y}. Let $\rchi$ denote a dataset of training images, where $\rchi_c \subset \rchi$ represents the subset that belongs to category $c \in {\bf Y}$, $\left(c = 1, 2, \ldots , C\right)$. We denote the score (logit) for class $c$ (before the softmax) as $y^c$ and the predicted class label as $c_{pred}$. Our high-level goal is to find a fault-line explanation ($\Psi$) that alters the CNN prediction from $c_{pred}$ to another specified class $c_{alt}$ using a minimal number of xconcepts. We follow \citep{kim2018interpretability} in defining the notion of xconcepts where each xconcept is represented using a set of example images. This representation of xconcepts provides great flexibility and portability as it will not be constrained to input features or a training dataset, and one can utilize the generated xconcepts across multiple datasets and tasks. 

We represent the quadruple $<$$I$, $c_{pred}$, $c_{alt}$$>$ as a human's query $Q$ that will be answered by showing a fault-line explanation $\Psi$. We use $\Sigma$ to represent all the xconcepts mined from $\rchi$. The xconcepts specific to the class $c_{pred}$ and $c_{alt}$ are represented as $\Sigma_{pred}$ and $\Sigma_{alt}$ respectively. Our strategy will be to first identify the xconcepts $\Sigma_{pred}$ and $\Sigma_{alt}$ and then generate a fault-line explanation by finding a minimal set of xconcepts from $\Sigma_{pred}$ and $\Sigma_{alt}$.
Formally, the objective is to find a fault-line that maximizes the posterior probability:

%\vspace{-2pt}
\begin{equation}
    \begin{aligned}
    \operatorname*{\arg\max}_{\Psi} P\left(\Psi, \Sigma_{pred}, \Sigma_{alt}, \Sigma \, \middle\vert \, Q\right)
    % \propto \,\, &\operatorname*{\arg\max}_{\Psi}  P\left(\Psi \, \middle\vert \, \Sigma_{pred}, \Sigma_{alt}, Q\right) \, P\left(\Sigma_{pred}, \Sigma_{alt}, \Sigma \, \middle\vert \, Q\right)
%     P\left(pg^{UinM}_i\mid h_i,q_i,T\right)\propto \\
% p\left(q_i \mid h_i,pg^{UinM}_i,T\right)p\left(h_i \mid pg^{UinM}_i,T\right)p\left(pg^{UinM}_i,T\right)
\end{aligned}
\end{equation}
% \begin{equation}
%     \begin{aligned}
    
\subsubsection{Mining Xconcepts}\label{mx}
 %As shown in Figure~\ref{fig:intro1}, (explain the xconcepts in the example )...
 We first compute $P\left(\Sigma \, \middle\vert \, \rchi, M\right)$ by identifying a set of semantically meaningful superpixels from every image and then perform clustering such that all the superpixels in a cluster are semantically similar. Each of these clusters represent an xconcept. We then identify class specific xconcepts i.e., $P\left(\Sigma_{pred} \, \middle\vert \, \Sigma , \rchi, I, c_{pred}, M\right)$ and $P\left(\Sigma_{alt} \, \middle\vert \, \Sigma , \rchi, I, c_{alt}, M\right)$. \\ \\
{A. Finding Semantically Meaningful Super-pixels as Xconcepts}\label{sp}

Figure~\ref{fig:model1} shows the overall algorithm for computing $P\left( \Sigma \, \middle\vert \, \rchi, M\right)$.
As deeper layers of the CNN capture richer semantic aspects of the image, we construct the xconcepts by making use of feature maps from the last convolution layer.  Let $f$ denote the feature extractor component of the CNN and $g$ denote the classifier component of the CNN that takes the output of $f$ and predicts log-probabilities over output classes ${\bf Y}$. We denote the $m$ feature maps produced at layer $L$ of the CNN as $A^{m,L} = \{a^L|a^L = f(I)\}$ which are of width $u$ and height $v$. We consider each feature map as an instance of an xconcept and obtain its localization map (i.e., super-pixels of each feature map). To produce the localization map, we use Grad-CAM~\citep{selvaraju2017grad} to compute the gradients of $y^c$ with respect to the feature maps $A^{m,L}$ and are then spatially pooled using Global Average Pooling (GAP) to obtain the importance weights ($\alpha_{m,L}^c$) of a feature map $m$ at layer $L$ for a target class $c$: 
\begin{equation}
    \begin{aligned}
    %   loss_{SH}(i, r) = \sum_{\hat{r}} [s(i, r) - s(i,\hat{r}) - \tau]_{+} + \sum_{\hat{i}} [s(i, r) - s(\hat{i},r) - \tau]_{+}
        \alpha_{m,L}^c = \frac{1}{Z}\sum_{i}\sum_{j}\frac{\partial y^c}{\partial A_{ij}^{m,L}}
    \end{aligned}
\end{equation}

Each element in the feature map $A^{m,L}$ is indexed by $i$, $j$ and $A_{ij}^{m,L}$ refers to the activation at location $(i,j)$ of the feature map $A^{m,L}$. $Z$ denotes the proportionality constant representing the total number of elements in $A^{m,L}$. Intuitively, Grad-CAM uses the gradient information flowing into the
last convolutional layer of the convolution network to assign importance values to each neuron. In other words, the gradients flowing back are global-average-pooled over the width and height dimensions to compute the importance weights $\alpha_{m,L}^c$.

Using the importance weights, we select top $p$ super-pixels for each class. Given that there are $C$ output classes in the dataset $\rchi$, we get $p*C$ super-pixels from each image in the training dataset. We apply K-means clustering with outlier removal to group these super-pixels into $G$ clusters where each cluster represents an xconcept (as shown in Figure~\ref{fig:model1}). For clustering, we consider the spatial feature maps $f(I)$ instead of the super-pixels (i.e., actual image regions) themselves. We use the silhouette score value of a different range of clusters to determine the value of K. \\ \\ 
{B. Identifying Class-Specific Xconcepts}\label{cs}
For each output class $c$, we learn the most common xconcepts that are highly influential in the prediction of that class over the entire training dataset $\rchi$. We use the TCAV technique~\citep{kim2018interpretability} to identify these class-specific xconcepts. Specifically, we construct a vector representation of each xconcept, called a CAV (denoted as $v_X$), by using a direction normal to a linear classifier trained to distinguish between the xconcept activations from the random activations. We then compute directional derivatives ($S_{c,X}$) to produce estimates of how important the concept $X$ was for a CNN's prediction of a target class $c$, e.g., how important the xconcept \texttt{stripedness} is for predicting the zebra class. 
\begin{equation}
    \begin{aligned}
        S_{c,X} = \nabla g_c (f(I)) \cdot v_X
    \end{aligned}
\end{equation}
where $g_c$ denote the classifier component of the CNN that takes the output of $f$ and predicts log-probability of output class $c$. Note that directional derivatives represent the derivative of logit values with respect to activations at the layer of interest, which helps in quantifying the model prediction’s sensitivity to a xconcept. We argue that these class-specific xconcepts facilitate in generating meaningful explanations by pruning out incoherent xconcepts. For example, the xconcepts such as \texttt{wheel} and \texttt{wings} are irrelevant in explaining why the network's prediction is a $zebra$ and not a $cat$. \\ \\
\subsubsection{Fault-Line Generation}\label{fl}
In this subsection, we describe our approach to generate a fault-line explanation using the class-specific xconcepts. 
Let us consider that $n_{pred}$ and $n_{alt}$ xconcepts have been identified for output classes $c_{pred}$ and $c_{alt}$ respectively, i.e., $\abs{\Sigma_{pred}} = n_{pred}$ and $\abs{\Sigma_{alt}} = n_{alt}$. We denote CAVs of the $n_{pred}$ xconcepts belonging to the class $c_{pred}$ as $v_{pred} = \{v_{pred}^{i},\, i = 1, 2, \ldots, n_{pred}\}$ and CAVs of the $n_{alt}$ xconcepts belonging to the class $c_{alt}$ as $v_{alt} = \{v_{alt}^{i}, \,i = 1, 2, \ldots, n_{alt}\}$. 
We formulate finding a fault-line explanation as the following optimization problem:
\begin{equation}
    \begin{aligned}
        &\underset{\delta_{pred},\delta_{alt}}{\text{minimize}} \;\;\;
       \alpha D(\delta_{pred}, \delta_{alt}) + \beta\left\|\delta_{pred}\right\|_1 + \lambda\left\|\delta_{alt}\right\|_1; \\
%     \end{aligned}
% \end{equation}
% where the distance function $D(I, I', \delta_{pred}, \delta_{alt})$ is defined as follows:
% \begin{equation}
%     \begin{aligned}
    &D(\delta_{pred}, \delta_{alt}) = max\{g^{pred}(I^{'}) - g^{alt}(I^{'}), -\tau\};\\
    &I^{'} = A^{m,L} \circ v_{pred}^\top\delta_{pred} \circ v_{alt}^\top\delta_{alt};\\
    &\delta_{pred}^i \in \{-1,0\},\;\delta_{alt}^i \in \{0,1\}  \; \forall i\; {\text{and}} \;\alpha, \beta,\lambda,\tau \geq 0. \\
\end{aligned}
\label{eq:obj}
\end{equation}

We elaborate on the role of each term in the Equation~\ref{eq:obj} as follows. Our goal here is to derive a fault-line explanation that gives us the minimal set of xconcepts from $\Sigma_{pred}$ and $\Sigma_{alt}$ that will alter the model prediction from $c_{pred}$ to $c_{alt}$.
Intuitively, we try creating new images ($I'$) by removing xconcepts in $\Sigma_{pred}$ from $I$ and adding xconcepts in $\Sigma_{alt}$ to $I$ until the classification result changes from $c_{pred}$ to $c_{alt}$. To do this, we do not directly perturb the original image but change the activations obtained at last convolutional layer $A^{m,L}$ instead. It may be noted that our goal is not to produce realistic images $I'$. We instead pick the most influential xconcepts by directly modifying the activation maps at a convolution layer\footnote{It is a very difficult task to produce realistic resulting images for datasets that have a diverse set of target classes.}.

In order to perturb the activations, we take the Hadamard product ($\circ$) between the activations ($A^{m,L}$), $v_{pred}^\top\delta_{pred}$ and $v_{alt}^\top\delta_{alt}$. The difference between the new logit scores for $c_{pred}$ (i.e., $g^{pred}(I')$) and $c_{alt}$ (i.e,. $g^{alt}(I')$) is controlled by the parameter $\tau$.

For any given confidence $\tau$ $>$ 0, the loss function $D(\delta_{pred}, \delta_{alt})$ is minimized when the new logit scores for $c_{pred}$ i.e., $g^{pred}(I')$ is smaller than logits for $c_{alt}$ i.e,. $g^{alt}(I')$ by at least $\tau$. Similar to \citep{dhurandhar2018explanations}, the terms $\beta\left\|\delta_{pred}\right\|_1$,  $\lambda\left\|\delta_{alt}\right\|_1$ in the optimization are introduced as $L_1$ regularizers to select sparse features. We apply a projected fast iterative shrinkage-thresholding algorithm (FISTA)~\citep{beck2009fast,dhurandhar2018explanations} for solving the above optimization problem. We outline our method in Algorithm~\ref{alg1}.
\begin{algorithm}
%\DontPrintSemicolon
  {input image $I$, classification model $M$, predicted class label $c_{pred}$, alternate class label $c_{alt}$ and training dataset $\rchi$}
 \begin{enumerate}[leftmargin=*,itemsep=0pt,topsep=0pt, partopsep=0pt]
\item Find semantically meaningful superpixels in $\rchi$,
 %\vspace{-1pt}
  \begin{equation}
  \nonumber
    \begin{aligned}
        \alpha_{m,L}^c = \frac{1}{Z}\sum_{i}\sum_{j}\frac{\partial y^c}{\partial A_{ij}^{m,L}}
    \end{aligned}
    %\vspace{-1pt}
\end{equation}
  \item Apply K-means clustering on superpixels and obtain xconcepts ($\Sigma$).
  \item Identify class specific xconcepts ($\Sigma_{pred}$ and $\Sigma_{alt}$) using TCAV,
  %\vspace{-1pt}
  \begin{equation}
  \nonumber
    \begin{aligned}
        S_{c,X} = \nabla g_c (f(I)) \cdot v_X
    \end{aligned}
%\vspace{-2pt}
\end{equation}
\item Solve Equation~\ref{eq:obj} to obtain fault-line $\Psi$,
%\vspace{-2pt}
\begin{equation}
\nonumber
    \begin{aligned}
     \Psi \leftarrow \underset{\delta_{pred},\delta_{alt}}{\text{min}}
       \alpha D(\delta_{pred}, \delta_{alt}) + \beta\left\|\delta_{pred}\right\|_1 + \lambda\left\|\delta_{alt}\right\|_1
\end{aligned}
%\vspace{-1pt}
\end{equation}
%\vspace{-5pt}
\end{enumerate}
\textbf{return} $\Psi$.
\caption{Generating Fault-Line Explanations}
\label{alg1}
\end{algorithm}

\begin{figure*}[t]
\centering
  \includegraphics[width=0.7\linewidth]{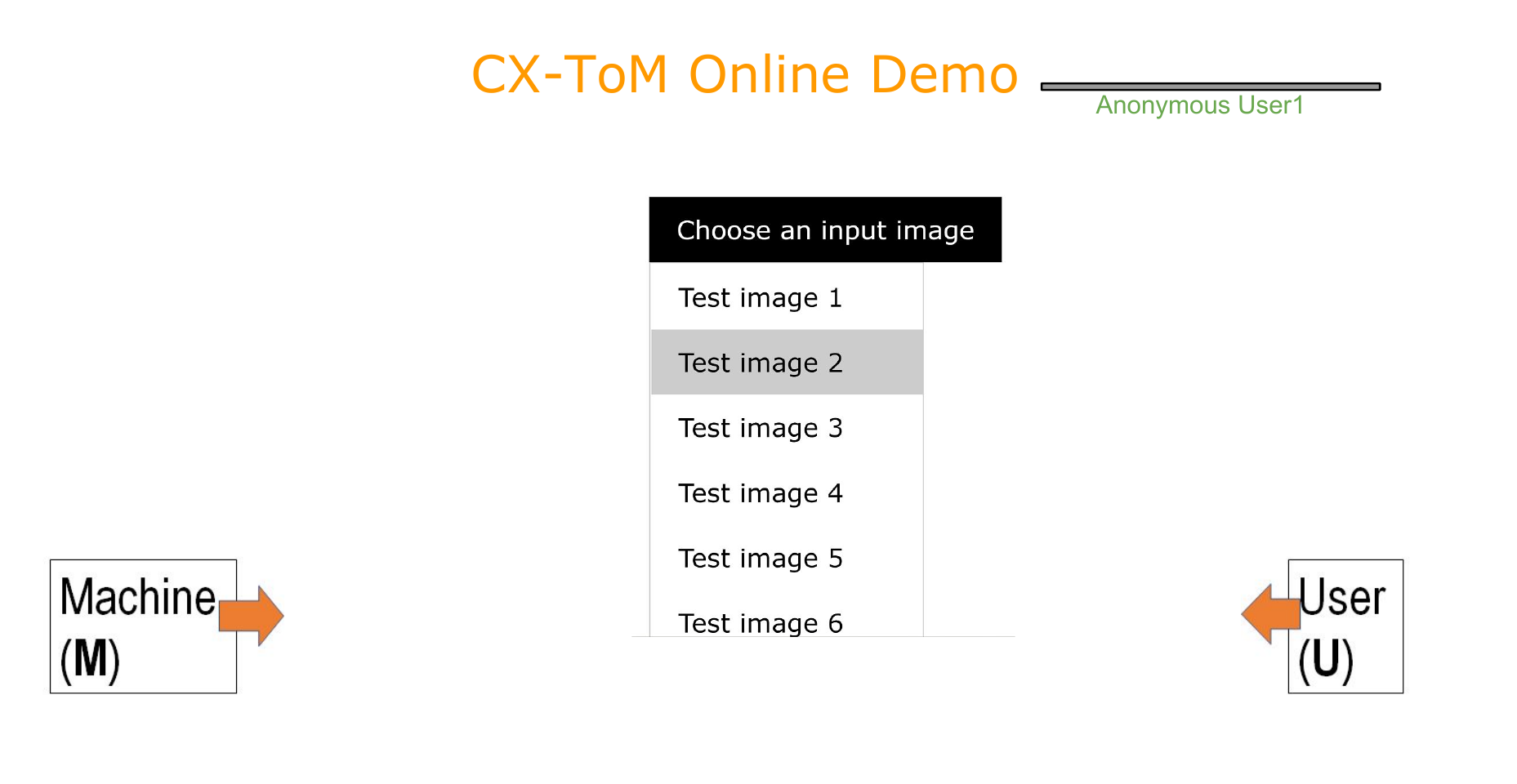}
  \caption{User interaction with CX-ToM in dialog to learn user preferences/utilities. User is first asked to select a input image.}~\label{fig:interaction_interface1}
  \vspace{-10pt}
\end{figure*}

\begin{figure*}[t]
\centering
  \includegraphics[width=0.8\linewidth]{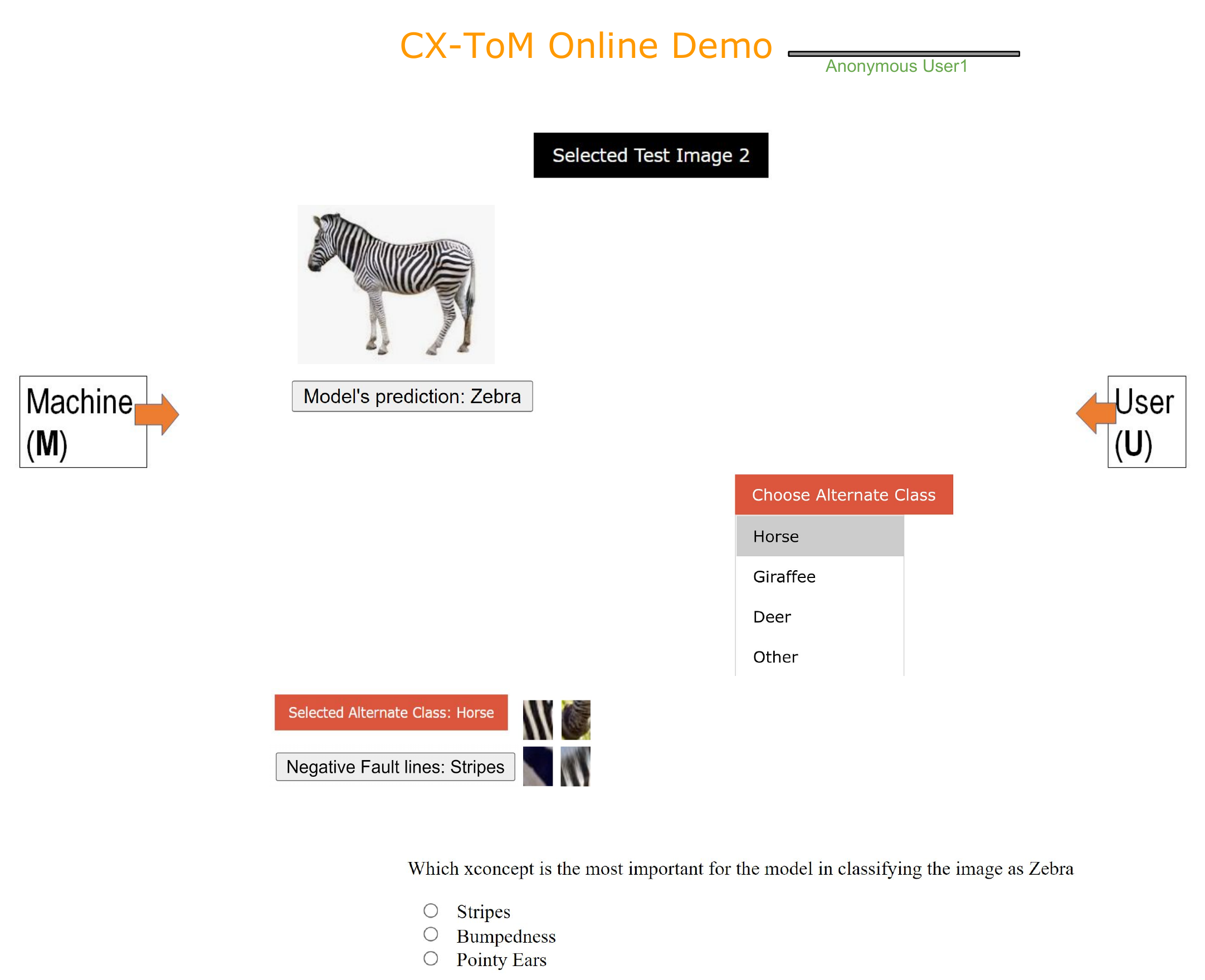}
  \caption{After the input image is selected by the user, user is then asked to select an alternate class to modify the model's decision. User is then shown a question to assess his/her understanding of model's important features in classifying the input image. If the user correctly answers the question, the reward is considered positive, otherwise negative.}~\label{fig:interaction_interface2}
  \vspace{-10pt}
\end{figure*}

\begin{table*}[t]
\small
  \centering
  \begin{tabular}{|P{0.3cm}|P{3.6cm}|P{2.2cm}|P{1.4cm}|P{1.4cm}|P{1.5cm}|P{2.4cm}|P{1.4cm}|}
    \toprule
     & \multicolumn{1}{P{2.7cm}|}{\vfil \hfil \textbf{XAI Framework}} & \multicolumn{1}{P{2cm}|}{\vfil \hfil \textbf{Justified Trust} ($\pm$std)} & \multicolumn{5}{P{8.5cm}|}{\vfil \hfil \textbf{Explanation Satisfaction} ($\pm$std)} \\ 
     \cline{4-8}
      & & & \vfil  Confidence  & \vfil \hfil Usefulness & \vfil Appropriate Detail & \vfil \hfil Understandability & \vfil  Sufficiency \\[0.3ex]
     \midrule
     \midrule
     \multirow{12}{*}{\STAB{\rotatebox[origin=c]{90}{\textbf{Non-Expert Subject Pool}}}}
     & Random Guessing & $\SI{6.6}{\percent}$ & N/A & N/A & N/A & N/A & N/A \\[0.3ex]
     
     & NO-X & $\SI{21.4 \pm 2.7}{\percent}$ & N/A & N/A & N/A & N/A & N/A \\[0.3ex]
     
     & CAM \citep{zhou2016learning} & $\SI{24.0 \pm 1.9}{\percent}$ & $4.2 \pm 1.8$ & $3.6 \pm 0.8$ & $2.2 \pm 1.9$ & $3.2 \pm 0.9$ & $2.6 \pm 1.3$ \\[0.3ex]
     
     & Grad-CAM \citep{selvaraju2017grad} & $\SI{29.2 \pm 3.1}{\percent}$ & $4.1 \pm 1.1$ & $3.2 \pm 1.9$ & $3.0 \pm 1.6$ & $4.2 \pm 1.1$ & $3.2 \pm 1.0$ \\
     & LIME \citep{ribeiro2016should} & $\SI{46.1 \pm 1.2}{\percent}$ & $5.1 \pm 1.8$ & $4.2 \pm 1.6$ & $3.9 \pm 1.1$ & $4.1 \pm 2.0$ & $4.3 \pm 1.6$ \\
     & SHAP \citep{NIPS2017_7062} & $\SI{40.9 \pm 2.0}{\percent}$ & $4.8 \pm 3.0$ & $3.9 \pm 1.1$ & $3.6 \pm 1.9$ & $3.8 \pm 1.4$ & $4.0 \pm 2.3$ \\
     & LRP \citep{bach2015pixel} & $\SI{31.1 \pm 2.5}{\percent}$ & $1.1 \pm 2.2$ & $2.8 \pm 1.0$ & $1.6 \pm 1.7$ & $2.8 \pm 1.0$ & $2.1 \pm 1.8$ \\[0.3ex]
     
     & SmoothGrad \citep{smilkov2017smoothgrad} & $\SI{37.6 \pm 2.9}{\percent}$ & $1.4 \pm 1.0$ & $2.2 \pm 1.8$ & $2.8 \pm 1.0$ & $3.1 \pm 0.8$ & $2.9 \pm 0.8$ \\
     & TCAV \citep{kim2018interpretability} & $\SI{49.7 \pm 3.3}{\percent}$ & $3.6 \pm 2.1$ & $3.2 \pm 1.8$ & $3.3 \pm 1.6$ & $3.6 \pm 2.1$ & $3.9 \pm 1.1$ \\[0.3ex]
    
     & CEM \citep{dhurandhar2018explanations} & $\SI{51.0 \pm 2.1}{\percent}$ & $4.1 \pm 1.4$ & $3.4 \pm 1.4$ & $3.1 \pm 2.1$ & $2.9 \pm 0.9$ & $3.3 \pm 1.6$ \\
     & CVE \citep{Goyal2019counterfactual} & $\SI{50.9 \pm 3.0}{\percent}$ & $3.8 \pm 1.9$ & $3.1 \pm 0.9$ & $3.6 \pm 2.1$ & $4.1 \pm 1.2$ & $4.2 \pm 1.2$ \\[0.3ex]
     
     & Fault-lines without ToM & $\SI{69.1 \pm 2.1}{\percent}$ & $6.2 \pm 1.2$ & $\mathbf{6.6} \pm \mathbf{0.7}$ & $7.2 \pm 0.9$ & $7.1 \pm 0.6$ & $6.2 \pm 0.8$ \\
     
    & CX-ToM (Fault-lines with ToM) & \textbf{$\SI{72.1 \pm 1.1}{\percent}$} & $\mathbf{6.9} \pm \mathbf{0.8}$ & $6.5 \pm 0.9$ & $\mathbf{7.8} \pm \mathbf{1.2}$ & $\mathbf{7.7} \pm \mathbf{0.2}$ & $\mathbf{6.9} \pm \mathbf{0.6}$ \\
    
    \midrule
    \midrule
    [0.3ex]
     \multirow{12}{*}{\STAB{\rotatebox[origin=c]{90}{\textbf{Expert Subject Pool}}}}
     & NO-X & $\SI{28.1 \pm 4.1}{\percent}$ & N/A & N/A & N/A & N/A & N/A \\[0.3ex]
     & CAM \citep{zhou2016learning} & $\SI{37.1 \pm 3.9}{\percent}$ & $3.2 \pm 1.8$ & $3.3 \pm 1.4$ & $3.1 \pm 2.1$ & $3.1 \pm 1.8$ & $2.9 \pm 1.9$ \\[0.3ex]
     & Grad-CAM \citep{selvaraju2017grad} & $\SI{39.1 \pm 2.1}{\percent}$ & $3.7 \pm 1.2$ & $3.1 \pm 2.2$ & $2.7 \pm 1.9$ & $3.7 \pm 1.1$ & $3.4 \pm 1.6$ \\
     & LIME \citep{ribeiro2016should} & $\SI{42.1 \pm 3.1}{\percent}$ & $3.1 \pm 2.2$ & $3.0 \pm 1.2$ & $2.8 \pm 1.9$ & $3.1 \pm 2.2$ & $2.8 \pm 1.7$ \\
     & LRP \citep{bach2015pixel} & $\SI{51.1 \pm 3.1}{\percent}$ & $3.2 \pm 4.1$ & $3.5 \pm 1.6$ & $4.2 \pm 1.5$ & $4.3 \pm 1.0$ & $3.9 \pm 0.9$ \\[0.3ex]
     & SmoothGrad \citep{smilkov2017smoothgrad} & $\SI{40.7 \pm 2.1}{\percent}$ & $3.1 \pm 1.0$ & $2.9 \pm 1.2$ & $3.8 \pm 1.5$ & $3.3 \pm 1.1$ & $3.1 \pm 1.0$ \\
     & TCAV \citep{kim2018interpretability} & $\SI{55.1 \pm 3.3}{\percent}$ & $3.9 \pm 2.8$ & $3.6 \pm 1.6$ & $4.1 \pm 1.3$ & $4.9 \pm 1.2$ & $3.9 \pm 0.8$ \\[0.3ex]
     & CEM \citep{dhurandhar2018explanations} & $\SI{61.1 \pm 2.2}{\percent}$ & $4.8 \pm 1.6$ & $3.7 \pm 1.6$ & $4.0 \pm 1.2$ & $3.7 \pm 1.0$ & $4.0 \pm 1.1$ \\
     & CVE \citep{Goyal2019counterfactual} & $\SI{64.5 \pm 3.7}{\percent}$ & $4.1 \pm 2.3$ & $3.9 \pm 1.5$ & $4.6 \pm 1.5$ & $4.5 \pm 1.4$ & $3.9 \pm 1.2$ \\[0.3ex]
     & Fault-lines without ToM & $\SI{70.5 \pm 1.3}{\percent}$ & $5.7 \pm 1.1$ & $4.9 \pm 0.8$ & $5.8 \pm 1.2$ & $6.9 \pm 1.1$ & $6.4 \pm 1.0$ \\
     & CX-ToM (Fault-lines with ToM) & \textbf{$\SI{74.5 \pm 0.7}{\percent}$} & $\mathbf{6.1} \pm \mathbf{0.8}$ & $\mathbf{5.3} \pm \mathbf{0.4}$ & $\mathbf{5.9} \pm \mathbf{1.2}$ & $\mathbf{7.1} \pm \mathbf{0.8}$ & $\mathbf{6.9} \pm \mathbf{0.7}$ \\
\bottomrule
  \end{tabular}
   \caption {Quantitative (Justified Trust) and Qualitative (Explanation Satisfaction) comparison of CX-ToM with random guessing baseline, no explanation (NO-X) baseline, and other state-of-the-art XAI frameworks such as CAM, Grad-CAM, LIME, LRP, SmoothGrad, TCAV, CEM, and CVE.}
    \label{tab1}
    %\vspace{-2pt}
\end{table*}

\subsection{CX-ToM Framework}%\label{fl}
In CX-ToM, we integrate both the ToM (Sec 3.1) and fault-lines (Sec 3.2) into one single explanation framework. Essentially, CX-ToM performs fault-line selection using ToM. Given an input image and two output categories, fault-lines  show  the  most  important  features  or  attributes  that influence  model’s  decision  in  classifying  the  image  as  one among the two output categories. In most cases, there exists several thousands of output categories and it is impossible for the human user to verify the model’s reasoning and behaviour by constructing a fault-line between all the possible pairs of output categories. Therefore, we learn an optimal policy to  automatically  select  the  most  important pair  for  constructing  fault-line  explanation  that  helps  human  user  to  quickly  understand  the  model’s  strengths  or weaknesses. This eliminates the need for the human user to see a large number of fault-lines before understanding the model's behaviour.

We cast this as a reinforcement learning (RL) problem where CX-ToM interacts with several human users in a dialog to learn user preferences/utilities that help them to understand the model in fewer dialogs (i.e. fewer number of fault-lines). We express reward in terms of a user feedback and the number of dialog turns (less the number of dialogs, higher is the reward). Figure~\ref{fig:interaction_interface1} and Figure~\ref{fig:interaction_interface2} show the user interaction interface.
In the interaction, user is first asked to select an image from a list of randomly drawn images from the training data (we only consider image classes for which we extracted xconcepts).  After the input image is selected, user is then asked to select an alternate class to which the model needs to modify its decision through fault-lines. We show the list of alternate classes through a dropdown list. The entries in this dropdown are dynamically loaded based on model's current state of the RL policy. CX-ToM shows the optimal fault-line to the user and tracks the sequence of user's preferences through the RL policy. After showing the fault-line, the CX-ToM assess user's understanding  of model's important features in classifying the input image. If the user correctly answers the question, the reward is considered positive, otherwise negative. The RL policy is updated after every 15 dialog interactions.

The RL policy is learned by a standard recurrent neural network, called Long-Short Term Memory (LSTM) \citep{hochreiter1997long}. In this paper we use a 2-layer LSTM parameterized by $\theta$. Thus, the goal of the policy learning is to estimate the LSTM parameters  $\theta$. We use actor-critic with experience replay for policy optimization~\citep{wang2016sample}. 
The training objective is to find  parametrised policy $\pi\left(a_i | s_i;\theta\right)$ that maximizes the expected reward $J(\theta)$ over all possible fault-line sequences given a starting state. The state of the RL policy ($s$) captures whether an image class is already selected in the dialog to generate a fault-line for the input image. Our goal is to learn the best user preferred alternate image classes for each prediction class. Similarly, the action space ($a$) constitute the set of all image classes. The gradient of the objective function has the following form:

\begin{equation}
%\[
\begin{aligned}
 \nabla_\theta J(\theta) = \mathbb{E}[\nabla_\theta\log\pi_\theta\left(a_i|s_i;\theta\right)A\left(s_i,a_i\right)]
\end{aligned}
\end{equation}
%\]

where $A\left(s_i,a_i\right) = Q\left(s_i,a_i\right) - V(s_i)$ is the advantage function~\citep{sutton2000policy}. $Q\left(s_i,a_i\right)$ is the standard Q-function, and $V(s_i)$ is the value (baseline) function aimed at reducing the variance of the estimated gradient. 
Intuitively, the above policy optimisation can be seen as the task of learning to select the sequence of responses (actions) at each turn which maximises the long-term objective defined by the reward function. The learning agent uses the value of the value function to update the optimal policy function. The policy function represents the probabilistic distribution of the action space. In other words, the learning agent determines the conditional probability that the agent chooses the action $a$ when in state $s$. We use the same specifications of $Q\left(s_i,a_i\right)$ and $V(s_i)$ as in \citep{sutton2000policy}. As in \citep{sutton2000policy}, we sample the dialog experiences randomly from the replay pool for training.

\begin{figure*}[t]
\centering
  \includegraphics[width=0.8\linewidth]{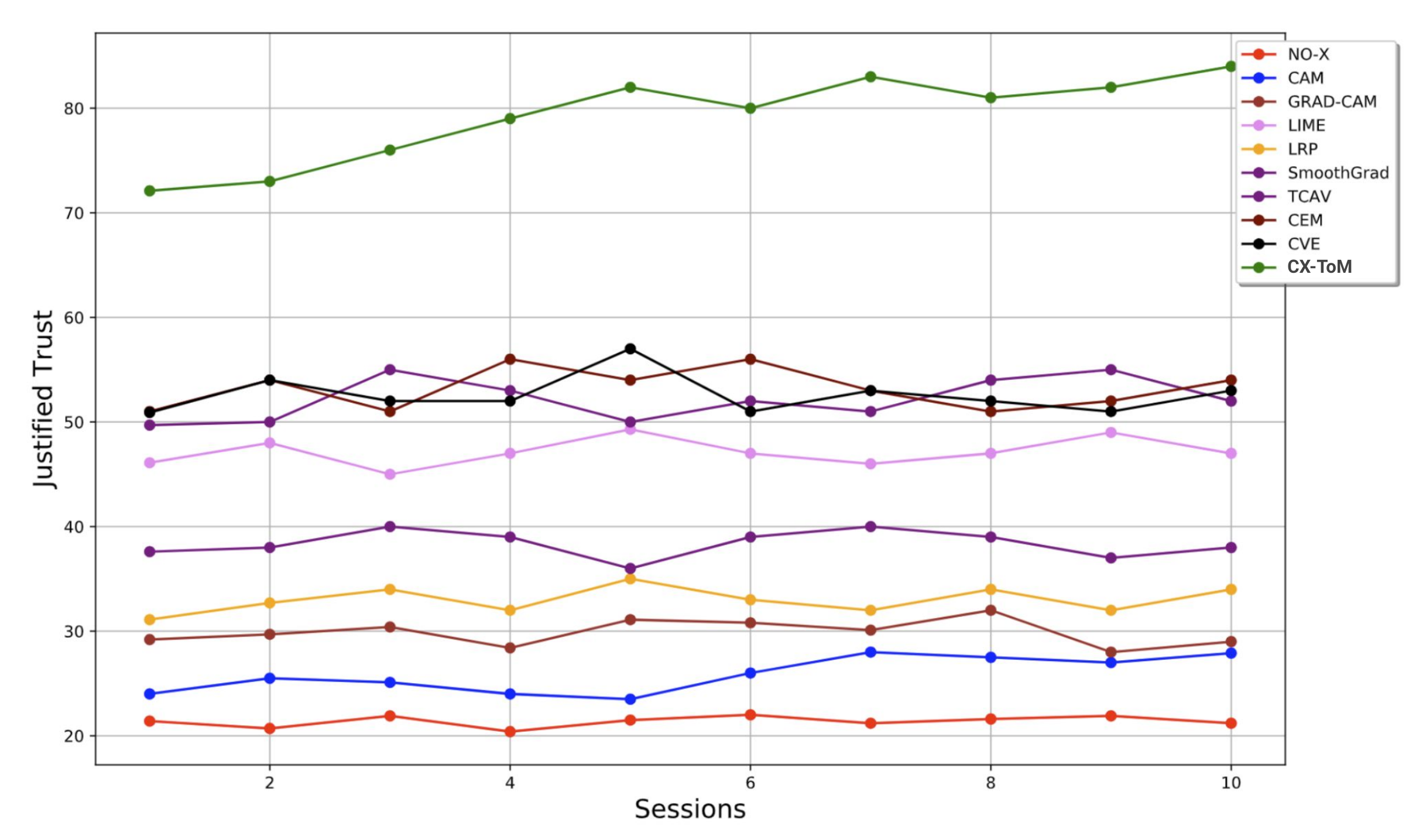}
  \caption{Gain in Justified Trust over time.}~\label{fig:exp1}
  %\vspace{-2pt}
\end{figure*}

\section{Experiments}
We conducted extensive human subject experiments to quantitatively and qualitatively assess the effectiveness of the proposed CX-ToM explanations in helping expert human users and non-expert human users understand the internal workings of the underlying model. We chose an image classification task for our experiments (although the proposed approach is generic and can be applied to any task). We use the following metrics~\citep{hoffman2018metrics,hoffman17explanation} to compare our method with the baselines\footnote{We empirically observed that the metrics Justified Trust and Explanation Satisfaction are effective in evaluating the core objective of XAI, i.e. to evaluate whether the user's understanding of the model improves with explanations. These metrics are originally defined at a high-level in the work by~\citep{hoffman2018metrics} and we adapt them for the image classification task.}.
\begin{enumerate}
\item \textbf{\textit{Justified Trust}} (Quantitative Metric). Justified Trust is computed by evaluating the human's understanding of the model's ($M$) decision-making process. In other words, given an image, it evaluates whether the users could reliably predict the model's output decision. More concretely, let us consider that $M$ predicts images in a set $C$ correctly and makes incorrect decisions on the images in the set $W$. Justified trust is given as sum of the percentage of images in $C$ that the human subject thinks $M$ would correctly predict and the percentage of images in $W$ that the human subject thinks $M$ would fail to predict correctly.
\item \textbf{\textit{Explanation Satisfaction (ES)}} (Qualitative Metric). We  measure  human  subjects’ feeling of satisfaction at having achieved an understanding of the machine in terms of usefulness, sufficiency, appropriated detail, confidence, and accuracy~\citep{hoffman2018metrics,hoffman17explanation}. We ask the subjects to rate each of these metrics on a Likert scale of 0 to 9. 
\end{enumerate}

We used ILSVRC2012 dataset (Imagenet)~\citep{russakovsky2015imagenet} and considered VGG-16~\citep{simonyan2014very} as the underlying network model. We randomly chose 80 classes in the dataset for our experiments and identified 57 xconcepts using our algorithm\footnote{We manually removed noisy xconcepts and fault-lines. We couldn't find an automatic approach to filter them. We leave this for future exploration.}.

We recruited 150 human subjects from our institution's Psychology subject pool~\footnote{These experiments were reviewed and approved by our institution's IRB.}. These subjects have no background in computer vision, deep learning or NLP (see Appendix) and we considered them as non-expert users. We recruited an additional 60 human subjects with background in computer vision. These subjects are experienced in training an image classification model using CNN and therefore we considered them as expert users. 

We applied between-subject design and randomly assigned subjects into eleven groups. We perform this separately with expert user pool and non-expert user pool. Each group in the non-expert pool are assigned 12 subjects and each group in the expert pool are assigned 5 subjects. Within each group, each subject will first go through a familiarization phase where the subjects become familiar with the underlying model through explanations (with 25 training images), followed by a testing phase where we apply our evaluation metrics and assess their understanding (on 8 test images) in the underlying model. We trained our ToM policy through the interactions with 15 subjects.
%Specifically, in the familiarization phase, human will be shown the input image $I$ and the CNN's prediction $c_{pred}$ and asked to provide $c_{alt}$ as input. We will then show an explanation to the human user for the model's prediction $c_{pred}$. For example, in CoCoX group, we show the fault-line explaining why the model chose $c_{pred}$ instead of $c_{alt}$. 
In the testing phase, human will be given only $I$ and will not see $c_{pred}$, $c_{alt}$, and explanations, and we evaluate whether the human can correctly identify $c_{pred}$ based on his/her understanding of the model gained in the familiarization phase. All our data and code will be made publicly available.

\begin{figure*}[t]
\centering
  \includegraphics[width=0.8\linewidth]{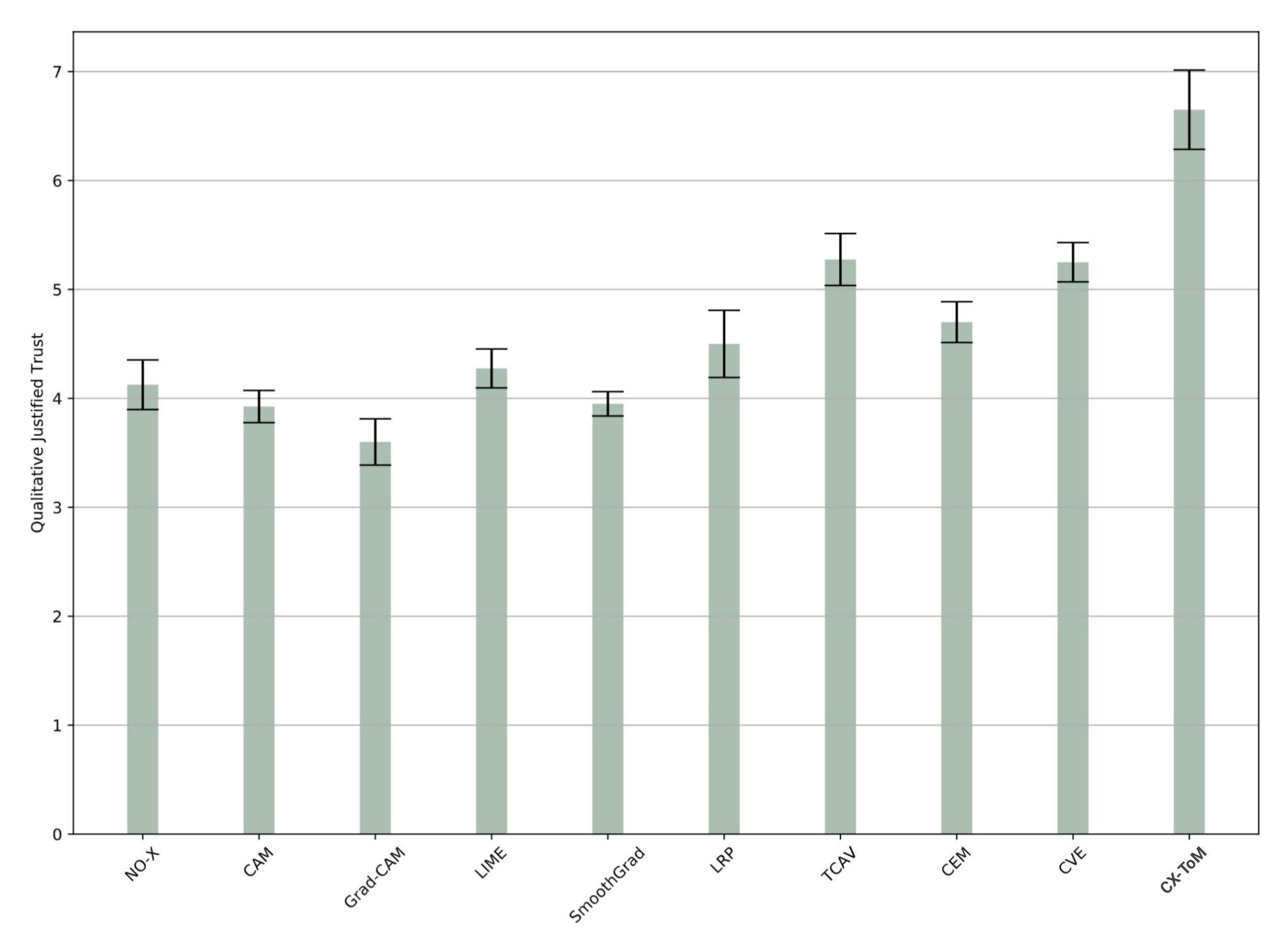}
  \caption{Average Qualitative Justified Trust (on a Likert scale of 0 to 9). Error bars denote standard errors of the means.}~\label{fig:subjective}
  %\vspace{-2pt}
\end{figure*}

For the first group, called NO-X (short for no-explanation group), we show the model's classification output on all the 25 images in the familiarization phase but we do not provide any explanation for the model's prediction. For the subjects in groups two to nine, in addition to the model's classification output, we also provide explanations in the familiarization phase for the model's prediction generated using the following state-of-the-art XAI models respectively: CAM~\citep{zhou2016learning}, Grad-CAM~\citep{selvaraju2017grad}, LIME~\citep{ribeiro2016should}, LRP~\citep{bach2015pixel}, SmoothGrad~\citep{smilkov2017smoothgrad}, TCAV~\citep{kim2018interpretability}, CEM~\citep{dhurandhar2018explanations}, and CVE~\citep{Goyal2019counterfactual}. For the subjects in the tenth group, we show the fault-line explanations without incorporating ToM policy. For the subjects in the eleventh group, we show the fault-line explanations selected based on our trained ToM policy. It may be noted that, in the testing phase, human will be shown only the image $I$ and will not be provided $c_{pred}$, $c_{alt}$, and explanations.

\subsection{Results}
Table~\ref{tab1} compares the Justified Trust (JT) and Explanation Satisfaction (ES) of all the groups in expert subject pool and non-expert subject pool. As we can see, JT and ES values of attention map based explanations such as Grad-CAM, CAM, and SmoothGrad do not differ significantly from the NO-X baseline, i.e., attention based explanations are not effective at increasing human trust and reliance (we did not evaluate ES for NO-X group as these subjects are not shown any explanations). This finding is consistent with the recent study by \citep{DBLP:journals/corr/abs-1902-10186} which shows that attention is not an explanation. 
On the other hand, concept based explanation framework TCAV and counterfactual explanation frameworks CEM, and CVE performed significantly better than the NO-X baseline (in both expert and non-expert pool). Our CX-ToM based explanations, which are both conceptual and counterfactual, significantly outperformed all the baselines. Note that, fault-lines with ToM policy performs better than randomly selecting a fault-line. Interestingly, expert users preferred LRP (JT = 51.1\%) to LIME (JT = 42.1\%) and non-expert users preferred LIME (JT = 46.1\%) to LRP (JT = 31.1\%).

\begin{figure*}[t]
\centering
  \includegraphics[width=\linewidth]{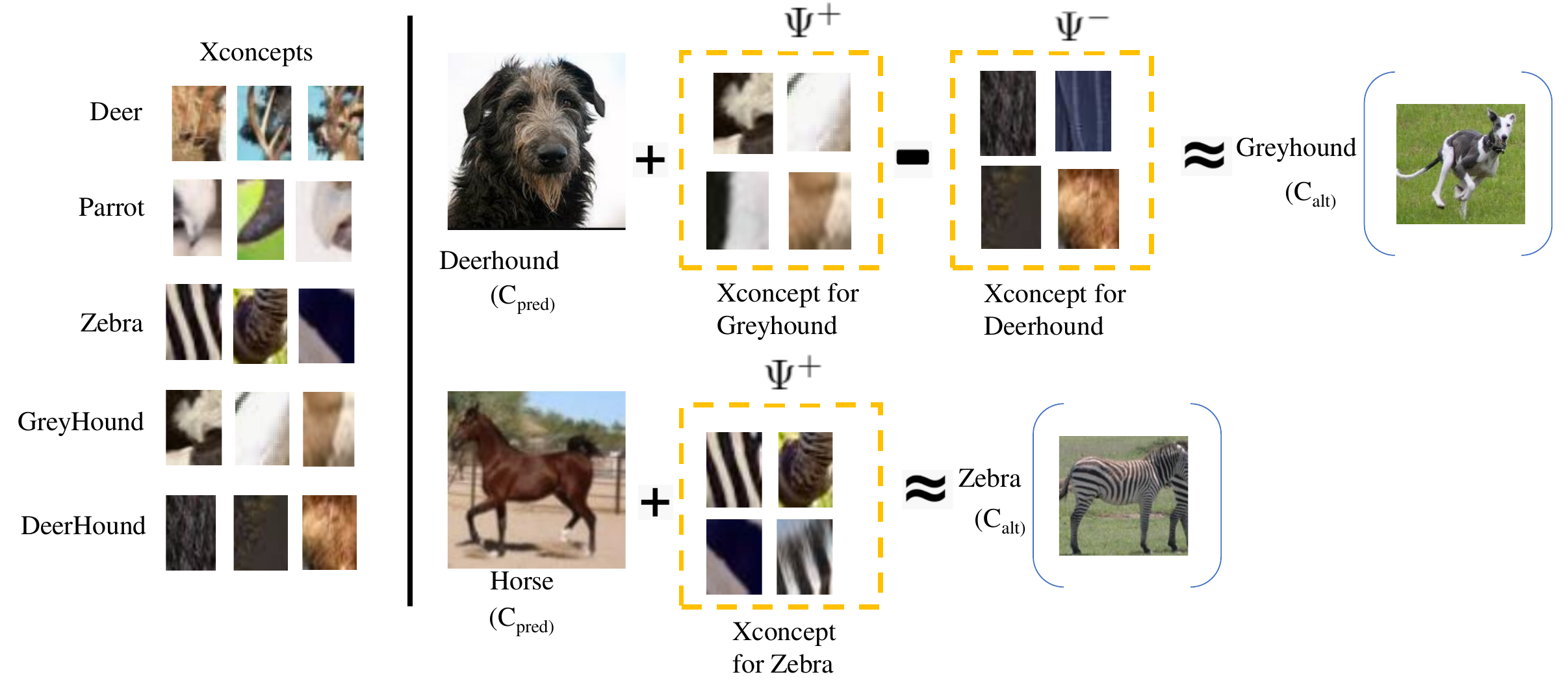}
  \caption{Examples of xconcepts (\textbf{Left}) and fault-line explanations (\textbf{Right}) identified by our method.}~\label{fig:cases}
  %\vspace{-2pt}
\end{figure*}

Furthermore, human subjects in our CX-ToM group, compared to all the other baselines, found that explanations are highly useful, sufficient, understandable, detailed and are more confident in answering the questions in the testing phase. These findings verify our hypothesis that fault-line explanations with ToM policy, are lucid and easy for both expert and non-expert users to understand~\footnote{Interestingly, we did not find significant differences across all the groups in terms of response time in answering the questions. We did an additional study with four subjects in each of the the groups to verify this and again found similar results. We leave this observation for future exploration.}. 

\textbf{Comparison with SHAP Baseline}:
We conduct an additional study to compare how the proposed method compares with SHAP approach~\citep{NIPS2017_7062}. SHAP, using shapley values, explains output predictions of a model for given input by computing the contribution of each feature to the prediction. Specifically, we use GradientExplainer~\footnote{\url{https://shap.readthedocs.io/en/latest/image\_examples.html}} implementation to compute SHAP explanations for the image classifier. We experiment with additional 12 human subjects (non-experts) to measure Justified Trust and Explanation Satisfaction. As shown in Table~\ref{tab1}, SHAP underperforms compared to CX-ToM and shows similar performance to LIME. This is expected as both LIME and SHAP are attribution based techniques.

\textbf{Gain in Justified Trust over Time}:
We hypothesized that subjects' justified trust in the CNN model might improve over time. This is because it can be harder for humans to fully understand the machine's underlying inference process in one single session. Therefore, we conduct an additional experiment with eight human subjects (non-experts) for each group where the subjects' reliance was measured after every session. Note that each session consists of a familiarization phase followed by a testing phase. The results are shown in Figure~\ref{fig:exp1}. As we can see, the subjects' JT in CX-ToM group increased at a higher-rate compared to other baselines. However, we did not find any significant increase in JT after fifth session across all the groups. This is consistent with our expectation that it is difficult for humans to focus on a task for longer periods~\footnote{In the future, we also intend to experiment with subjects by arranging sessions over days or weeks instead of having continuous back to back sessions.}. It should be noted that the increase in JT with attention map based explanations such as Grad-CAM and CAM is not significant. This finding again demonstrates that attention maps are not effective to improve human trust.

\textbf{Subjective Evaluation of Justified Trust}: In addition to the quantitative evaluation of the justified trust, we also collect subjective trust values (on a Likert scale of 0 to 9) from the subjects. This helps in understanding to what extent the users think they trust the model. The results are shown in Figure~\ref{fig:subjective}. As we can see, these results are consistent with our quantitative trust measures except that qualitative trust in Grad-CAM, CAM, and SmoothGrad is lower compared to the NO-X group.

\textbf{Case Study}:
Figure~\ref{fig:cases} shows examples of the xconcepts (cropped and rescaled for better view) identified using our approach. As we can see, our method successfully extracts semantically coherent xconcepts such as \textit{pointed curves} of \texttt{deer}, \textit{stripedness} of \texttt{zebra}, and \textit{woolliness} of \texttt{deerhound} from the training dataset. Also the fault-lines generated by our method correctly identify the most critical xconcepts that can alter the classification result from $c_{pred}$ to $c_{alt}$. For example, consider the image of \texttt{deerhound} shown in the Figure~\ref{fig:cases}. Our fault-line explanation suggests removing \textit{woolliness} and adding \textit{black and white pattern} to alter the model's classification on the image from \texttt{deerhound} to \texttt{greyhound}.

\begin{table*}[t]
\small
  \centering
  \begin{tabular}{|P{0.3cm}|P{3.6cm}|P{2.2cm}|P{1.4cm}|P{1.4cm}|P{1.5cm}|P{2.4cm}|P{1.4cm}|}
    \toprule
     & \multicolumn{1}{P{2.7cm}|}{\vfil \hfil \textbf{XAI Framework}} & \multicolumn{1}{P{2cm}|}{\vfil \hfil \textbf{Justified Trust} ($\pm$std)} & \multicolumn{5}{P{8.5cm}|}{\vfil \hfil \textbf{Explanation Satisfaction} ($\pm$std)} \\ 
     \cline{4-8}
      & & & \vfil  Confidence  & \vfil \hfil Usefulness & \vfil Appropriate Detail & \vfil \hfil Understandability & \vfil  Sufficiency \\[0.3ex]
     \midrule
     \midrule
     \multirow{6}{*}{\STAB{\rotatebox[origin=c]{90}{\textbf{Non-Expert Pool}}}}
     & Grad-CAM \citep{selvaraju2017grad} & $\SI{21.6 \pm 2.8}{\percent}$ & $3.2 \pm 1.5$ & $3.2 \pm 1.6$ & $2.7 \pm 2.8$ & $3.0 \pm 2.0$ & $2.9 \pm 0.9$ \\
     & LIME \citep{ribeiro2016should} & $\SI{26.9 \pm 3.5}{\percent}$ & $3.3 \pm 2.5$ & $3.1 \pm 2.1$ & $3.7 \pm 1.9$ & $3.1 \pm 1.8$ & $4.0 \pm 1.3$ \\
     & TCAV \citep{kim2018interpretability} & $\SI{42.2 \pm 2.6}{\percent}$ & $4.1 \pm 2.7$ & $3.2 \pm 2.4$ & $3.8 \pm 1.9$ & $4.0 \pm 1.5$ & $3.5 \pm 1.8$ \\[0.3ex]
    
     & CVE \citep{Goyal2019counterfactual} & $\SI{38.1 \pm 3.5}{\percent}$ & $2.7 \pm 2.5$ & $2.6 \pm 1.5$ & $3.0 \pm 2.0$ & $3.2 \pm 1.1$ & $3.2 \pm 1.9$ \\[0.3ex]
     
     & Fault-lines without ToM & $\SI{54.2 \pm 2.4}{\percent}$ & $6.1 \pm 1.7$ & $5.9 \pm 1.2$ & $6.6 \pm 1.5$ & $6.4 \pm 0.9$ & $6.2 \pm 1.1$ \\
     
    & CX-ToM (Fault-lines with ToM) & \textbf{$\SI{58.3 \pm 1.8}{\percent}$} & $\mathbf{6.3} \pm \mathbf{1.8}$ & $\mathbf{6.2} \pm \mathbf{1.6}$ & $\mathbf{6.9} \pm \mathbf{1.1}$ & $\mathbf{7.2} \pm \mathbf{0.8}$ & $\mathbf{7.2} \pm \mathbf{1.6}$ \\
    
    \midrule
    \midrule
    [0.3ex]
     \multirow{6}{*}{\STAB{\rotatebox[origin=c]{90}{\textbf{Expert Pool}}}}
     & Grad-CAM \citep{selvaraju2017grad} & $\SI{20.1 \pm 1.8}{\percent}$ & $2.5 \pm 2.2$ & $2.5 \pm 1.8$ & $1.7 \pm 1.9$ & $3.0 \pm 1.9$ & $3.0 \pm 1.2$ \\
     & LIME \citep{ribeiro2016should} & $\SI{25.4 \pm 2.7}{\percent}$ & $3.0 \pm 1.6$ & $3.2 \pm 2.9$ & $3.8 \pm 2.1$ & $2.6 \pm 1.0$ & $2.5 \pm 2.9$ \\
     & TCAV \citep{kim2018interpretability} & $\SI{46.0 \pm 2.4}{\percent}$ & $3.5 \pm 1.4$ & $3.8 \pm 1.7$ & $3.6 \pm 2.2$ & $3.8 \pm 2.1$ & $4.0 \pm 1.9$ \\[0.3ex]
     & CVE \citep{Goyal2019counterfactual} & $\SI{43.1 \pm 3.1}{\percent}$ & $3.2 \pm 2.3$ & $3.2 \pm 0.9$ & $3.0 \pm 1.8$ & $3.0 \pm 1.3$ & $3.4 \pm 1.8$ \\[0.3ex]
     & Fault-lines without ToM & $\SI{54.9 \pm 1.6}{\percent}$ & $\mathbf{6.2} \pm \mathbf{2.1}$ & $6.0 \pm 1.2$ & $5.3 \pm 1.6$ & $6.0 \pm 1.5$ & $5.9 \pm 1.5$ \\
     & CX-ToM (Fault-lines with ToM) & \textbf{$\SI{56.0 \pm 1.9}{\percent}$} & $5.8 \pm 1.6$ & $\mathbf{6.1} \pm \mathbf{1.0}$ & $\mathbf{6.1} \pm \mathbf{1.0}$ & $\mathbf{7.0} \pm \mathbf{1.5}$ & $\mathbf{7.0} \pm \mathbf{1.2}$ \\
\bottomrule
  \end{tabular}
   \caption {Justified Trust and Explanation Satisfaction Results of CX-ToM and baselines on ResNet-50.}
    \label{tab2}
    \vspace{-2pt}
\end{table*}

\begin{figure}[t]
\centering
  \includegraphics[width=0.99\linewidth]{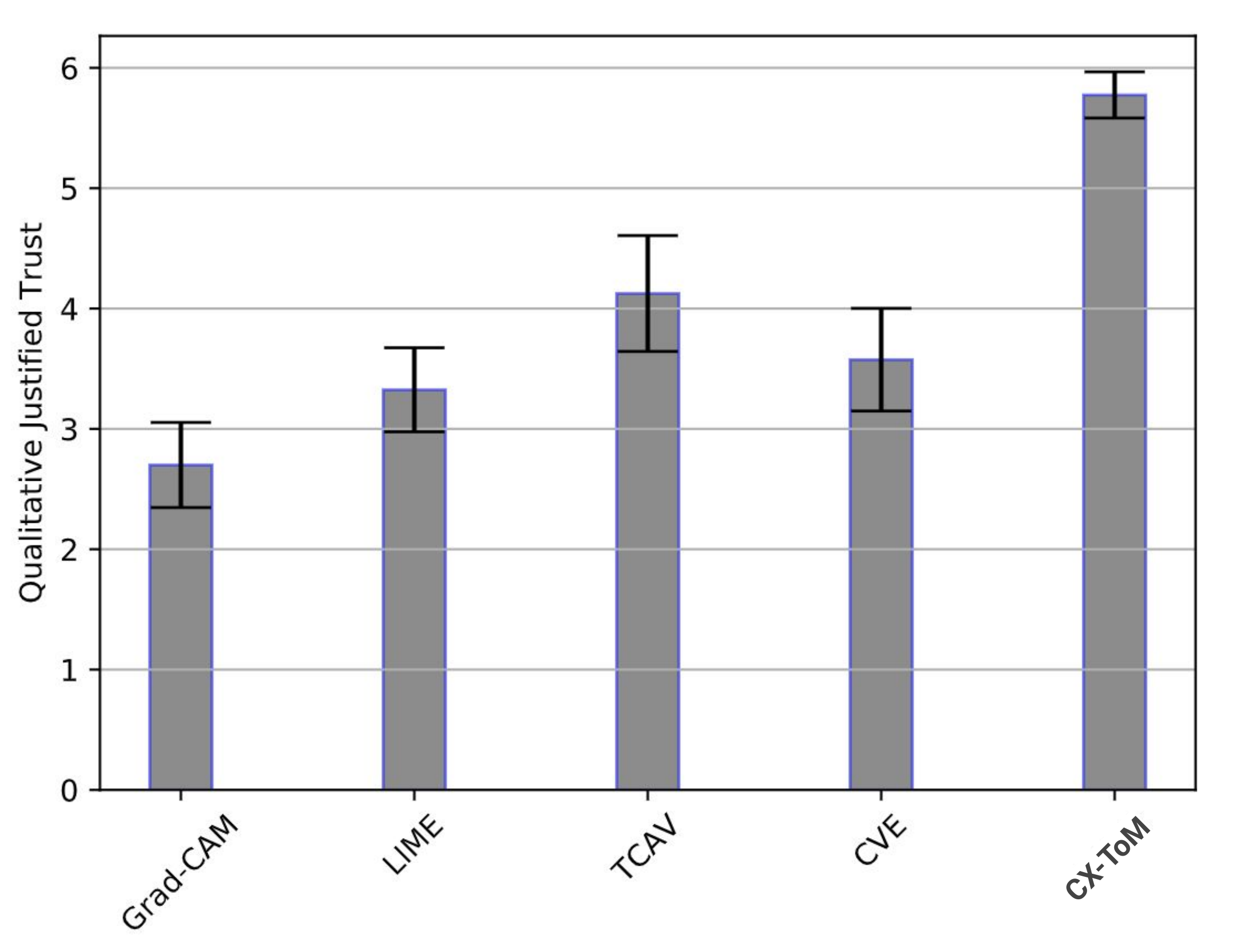}
  \caption{Average Subjective Justified Trust (on a Likert scale of 0 to 9) on ResNet-50.}~\label{fig:subjective2}
  \vspace{-2pt}
\end{figure}

\subsection{Additional Experiments}
In this section, we further assess the effectiveness of the proposed CX-ToM framework using more diverse and recent models as the underlying convolution neural network. 

\noindent
\textbf{ResNet50}:
ResNet~\citep{he2016deep} is a relatively deeper convolution neural network than VGG-16. It incorporates skip connections and batch normalization which greatly improves model's generalization capability and performance. More specifically, each ResNet block is 3 layer deep consisting of $1 \times 1$, $3 \times 3$, $1 \times 1$ convolutions respectively. The $1\times 1$ convolution layers are useful in reducing and then restoring the dimensions. Finally, the average pooling is performed and end it with a fully connected layer.

We apply our CX-ToM framework to ResNet. As discussed in Section 3.2 and Algorithm 1, we mine xconcepts from ResNet by producing localization maps. The average pooling layer is used to obtain importance weights of a feature map at a layer $L$ for a given target class. We obtain class-specific xconcepts using concept activation vectors. Finally fault-lines are generated by solving the optimization problem in equation (5).

% 1 para on experiment set up and the metrics, dataset and the baselines
We use ILSVRC2012 dataset for our experiments. We compare our approach against the following baselines: Grad-CAM~\citep{selvaraju2017grad}, LIME~\citep{ribeiro2016should}, TCAV~\citep{kim2018interpretability}, and CVE~\citep{Goyal2019counterfactual}. Similar to our experiments with VGG-16, we use the metrics Justified Trust (JT) and Explanation Satisfaction (ES) to compare our approach with baselines. We recruited human subjects from our institution's Psychology subject pool. We apply between-subject design and randomly assigned subjects into six groups. Each group in the non-expert pool are assigned 4 subjects and each group in the expert pool are assigned 2 subjects. We have identified 15 xconcepts and closely followed the experiment setup and design used in our experiments on VGG-16 model. 

Table~\ref{tab2} summarizes the JT and ES results of all the six groups. Similar to the results with VGG-16, trust improvements with Grad-CAM on both expert and non-expert pool is the least compared to other baselines. Among the baselines, TCAV is the best performing model, implying that concept level explanations are relatively more scalable to deeper networks than attention based explanations. Our CX-ToM based framework show significant improvements over TCAV baseline. The subjective evaluation of JT and ES  shows in Figure~\ref{fig:subjective2} further validate our hypotheses.

\begin{table*}[t]
\small
    \centering
  \begin{tabular}{|P{0.3cm}|P{3.6cm}|P{2.2cm}|P{1.4cm}|P{1.4cm}|P{1.5cm}|P{2.4cm}|P{1.4cm}|}
    \toprule
     & \multicolumn{1}{P{2.7cm}|}{\vfil \hfil \textbf{XAI Framework}} & \multicolumn{1}{P{2cm}|}{\vfil \hfil \textbf{Justified Trust} ($\pm$std)} & \multicolumn{5}{P{8.5cm}|}{\vfil \hfil \textbf{Explanation Satisfaction} ($\pm$std)} \\ 
     \cline{4-8}
      & & & \vfil  Confidence  & \vfil \hfil Usefulness & \vfil Appropriate Detail & \vfil \hfil Understandability & \vfil  Sufficiency \\[0.3ex]
     \midrule
     \midrule
     \multirow{6}{*}{\STAB{\rotatebox[origin=c]{90}{\textbf{Non-Expert Pool}}}}
     & Grad-CAM \citep{selvaraju2017grad} & $\SI{15.2 \pm 1.5}{\percent}$ & $2.4 \pm 1.8$ & $2.6 \pm 1.2$ & $2.5 \pm 1.5$ & $2.9 \pm 1.7$ & $3.0 \pm 1.2$ \\
     & LIME \citep{ribeiro2016should} & $\SI{22.2 \pm 2.4}{\percent}$ & $3.1 \pm 2.2$ & $2.7 \pm 2.0$ & $3.5 \pm 1.9$ & $2.7 \pm 1.2$ & $3.8 \pm 1.6$ \\
     & TCAV \citep{kim2018interpretability} & $\SI{40.1 \pm 2.2}{\percent}$ & $3.9 \pm 1.7$ & $3.6 \pm 1.1$ & $4.1 \pm 2.5$ & $4.0 \pm 1.2$ & $3.6 \pm 1.8$ \\[0.3ex]
    
     & CVE \citep{Goyal2019counterfactual} & $\SI{41.5 \pm 3.2}{\percent}$ & $3.1 \pm 1.5$ & $3.3 \pm 1.0$ & $3.8 \pm 2.1$ & $3.8 \pm 2.0$ & $3.9 \pm 1.2$ \\[0.3ex]
     
     & Fault-lines without ToM & $\SI{53.8 \pm 1.9}{\percent}$ & $\mathbf{6.3} \pm \mathbf{2.0}$ & $5.6 \pm 1.1$ & $6.1 \pm 1.9$ & $5.9 \pm 0.6$ & $6.6 \pm 1.6$ \\
     
    & CX-ToM (Fault-lines with ToM) & \textbf{$\SI{54.8 \pm 2.0}{\percent}$} & $6.2 \pm 2.0$ & $\mathbf{6.5} \pm \mathbf{1.8}$ & $\mathbf{6.2} \pm \mathbf{1.0}$ & $\mathbf{7.0} \pm \mathbf{1.9}$ & $\mathbf{6.8} \pm \mathbf{1.9}$ \\
    
    \midrule
    \midrule
    [0.3ex]
     \multirow{6}{*}{\STAB{\rotatebox[origin=c]{90}{\textbf{Expert Pool}}}}
     & Grad-CAM \citep{selvaraju2017grad} & $\SI{16.8 \pm 1.9}{\percent}$ & $2.3 \pm 1.2$ & $2.9 \pm 1.4$ & $2.0 \pm 1.9$ & $3.1 \pm 1.5$ & $3.2 \pm 2.2$ \\
     & LIME \citep{ribeiro2016should} & $\SI{23.7 \pm 2.0}{\percent}$ & $2.9 \pm 1.3$ & $3.2 \pm 2.5$ & $3.0 \pm 2.1$ & $2.5 \pm 1.6$ & $2.9 \pm 2.0$ \\
     & TCAV \citep{kim2018interpretability} & $\SI{38.6 \pm 3.1}{\percent}$ & $3.9 \pm 1.3$ & $3.2 \pm 1.5$ & $3.9 \pm 2.0$ & $4.0 \pm 1.0$ & $3.7 \pm 1.1$ \\[0.3ex]
     & CVE \citep{Goyal2019counterfactual} & $\SI{39.1 \pm 2.0}{\percent}$ & $3.5 \pm 2.2$ & $3.7 \pm 1.6$ & $3.2 \pm 1.2$ & $3.9 \pm 1.1$ & $3.0 \pm 1.5$ \\[0.3ex]
     & Fault-lines without ToM & $\SI{57.0 \pm 1.8}{\percent}$ & $6.0 \pm 1.5$ & $6.2 \pm 1.7$ & $5.8 \pm 1.9$ & $5.5 \pm 1.1$ & $6.1 \pm 1.9$ \\
     & CX-ToM (Fault-lines with ToM) & \textbf{$\SI{59.8 \pm 1.6}{\percent}$} & $\mathbf{6.3} \pm \mathbf{1.1}$ & $\mathbf{6.5} \pm \mathbf{1.7}$ & $\mathbf{7.0} \pm \mathbf{1.5}$ & $\mathbf{6.7} \pm \mathbf{1.7}$ & $\mathbf{6.5} \pm \mathbf{1.0}$ \\
\bottomrule
  \end{tabular}
   \caption {Justified Trust and Explanation Satisfaction Results of CX-ToM and baselines on PACNet.}
    \label{tab3}
    %\vspace{-2pt}
\end{table*}

\begin{figure}[t]
\centering
  \includegraphics[width=0.99\linewidth]{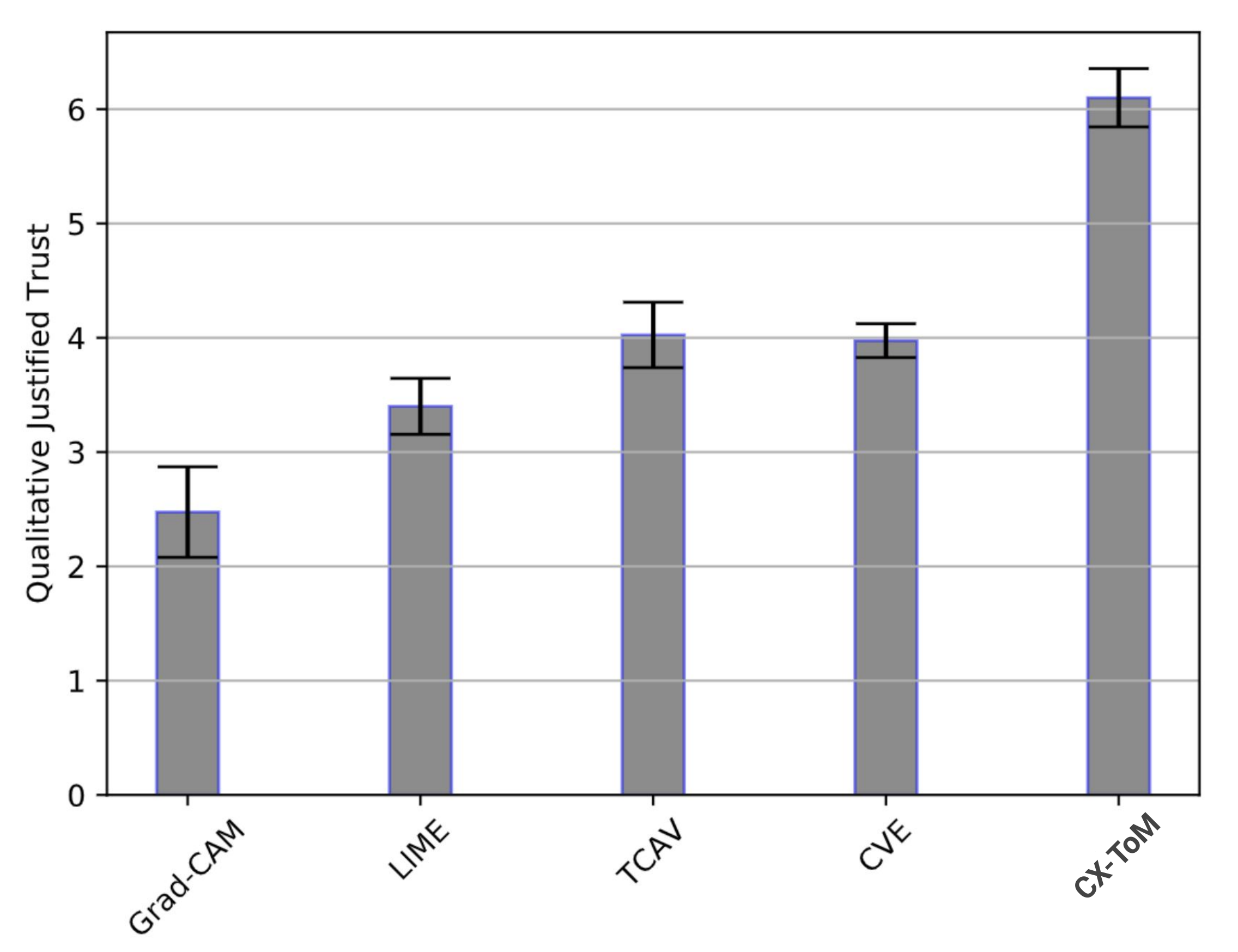}
  \caption{Average Subjective Justified Trust (on a Likert scale of 0 to 9) on PACNet.}~\label{fig:subjective3}
  %\vspace{-2pt}
\end{figure}

\noindent
\textbf{PAC Networks:}
Recently a pixel-adaptive convolution network called PAC~\citep{su2019pixel} is proposed to address the content-agnostic limitations of traditional CNNs. Specifically, in traditional CNNs, same convolutional filter banks are applied to all the input images irrespective of their content. However, image content varies substantially across the input images and therefore applying content-agnostic filter banks may not be optimal for all image types as well as different pixels in an image. In PAC networks, content-adaptive convolution operations are performed where a standard spatially invariant convolution filter $W$ is multiplied with a adapting kernel $K$. These networks are shown to be effective in a wide range of computer vision problems such as depth and optical flow upsampling tasks~\citep{su2019pixel}. 

We apply our CX-ToM framework to PACNet. Using Algorithm 1, we extract xconcepts from PACNet and obtain class-specific xconcepts using concept activation vectors. Finally fault-lines are generated by solving the optimization problem in equation (5). We use ILSVRC2012 dataset and consider the following baselines: Grad-CAM~\citep{selvaraju2017grad}, LIME~\citep{ribeiro2016should}, TCAV~\citep{kim2018interpretability}, and CVE~\citep{Goyal2019counterfactual}. We use the metrics Justified Trust (JT) and Explanation Satisfaction (ES) to compare our approach with baselines. We recruited human subjects from our institution's Psychology subject pool. We apply between-subject design and randomly assigned subjects into six groups. Each group in the non-expert pool are assigned 4 subjects and each group in the expert pool are assigned 2 subjects. We have identified 18 xconcepts and closely followed the experiment setup and design used in our experiments on VGG-16 and ResNet models.   

We present the JT and ES of all the six groups in Table~\ref{tab3}. As we can see, trust improvements with Grad-CAM on PACNet is relatively lower compared to VGG-16 and ResNet. This indicates that attention based explanations need more fine-tuning on the non-traditional CNN architectures. TCAV and CVE clearly outperform other baselines. Our CX-ToM based framework show relatively significant improvements over all the baselines indicating that our approach generalizes well to the recent CNN models. The subjective evaluation results of JT and ES shown in Figure~\ref{fig:subjective3} are consistent with our quantitative results.

\noindent
\textbf{Competency Testing}:
We perform competency testing experiment where we train two different CNNs, namely; AlexNet and ResNet-50. It may be noted that ResNet-50 is known to be more reliable and accurate than AlexNet. We show the predictions and the explanations from each of the two networks to the subjects (4 subjects in each of the above groups) and ask them to compare the reliability (competency) of the models relative to each other. We record the subjects' confidence scores in their answers on a Likert scale of 0 to 9. We chose only those images for both models made the same prediction as ground truth. The assumption here is that an effective and useful explanation helps the subject to distinguish between a reliable model and an unreliable model easily. We find that human subjects, who are shown CX-ToM explanations, are able to identify the more accurate and reliable classifier (i.e., ResNet-50) with high confidence (average confidence score = 7.7). Human subjects who are shown explanations based on Grad-CAM, CEM, and TCAV also identified that ResNet-50 is more reliable than AlexNet. However, they are not confident in their answers (avg. confidence scores are 2.6 (Grad-CAM), 4.9 (TCAV), and 4.2 (CEM)). Subjects in the remaining groups failed to identify the more reliable classifier.

\noindent
\textbf{Computational Cost}:
We run all components of our framework on one RTX 2080ti GPU. The extraction of super-pixels using Grad-CAM, discussed in section 3.2, takes about 17 hours (15 minutes per 100 images in the training dataset). The clustering of these super-pixels is relatively fast and completes within 3 hours to extract the 57 xconcepts from 80 image classes. Using TCAV technique to learn CAVs takes about 15 hours on RTX 2080ti and then identifying the directional derivatives takes about 2 hours for the extracted 57 xconcepts (discussed in section 3.2). Finally, the optimization step to select the appropriate fault-line takes about 40 seconds per image.

\section{Conclusions}
In this paper, we introduced a new explainable AI (XAI) framework, CX-ToM, based on Theory of Mind and fault-lines. We argue that due to their iterative, conceptual and counterfactual nature, CX-ToM based explanations are lucid, clear and easy for humans to understand. We proposed a new method to automatically mine explainable concepts from a given training dataset and to derive fault-line explanations. Moreover, we show that estimating the human’s understanding of the CNN model using Theory-of-Mind help in generating more appropriate fault-lines. Using qualitative and quantitative evaluation metrics, we demonstrated that CX-ToM significantly outperform baselines in improving human understanding of the underlying classification model.

\section*{STAR Methods}

\subsection*{RESOURCE AVAILABILITY}
\noindent
\textbf{Lead Contact:}\\
Further information and requests for resources and reagents should be directed to and will be fulfilled by the Lead Contact (Full Name: Arjun Reddy Akula; Email Address: aakula@ucla.edu) \\

\noindent
\textbf{Materials Availability:}\\
This study did not generate new unique reagents.\\

\noindent
\textbf{Data and Code Availability:}\\
The code is available publicly at this github page: \url{https://github.com/arjunakula/faultline\_explainer}

\subsection*{METHOD DETAILS}
In our human study experiments, we recruited 120 human subjects from our institution's Psychology subject pool. These experiments were reviewed and approved by our institution's IRB. We applied between-subject design and randomly assigned each subject into one of the experiment and control groups. We did not leverage any dataset from other publications. We leveraged the TCAV~\cite{kim2018interpretability} code to generate explainable concepts.

\subsection*{ADDITIONAL RESOURCES}
Our study has not generated or contributed to a new website.

\subsection*{KEY RESOURCES TABLE}
We attached the KRT along with this email (Reference Id: KRT61a93f53e38a6). Please see the file KRT61a93f53e38a6.docx.

\section*{Author Contributions}
A.A is the main lead author of the work. A.A implemented the explanation learning framework and also designed, conducted human study experiments, and wrote the paper. K.W. provided feedback in the initial stages of designing the learning framework. He is also a key part of the experiment design section. C.L, S.S, and H.L. provided  valuable feedback in improving the clarity of the paper and also helped in designing and conducting human studies. S.T., J.C. and S.C. played a major role in shaping up the overall explanation framework and also helped in designing experiments and in writing the work.

\section*{Acknowledgments}
The authors thank  Zhao Weng (UCLA), Sparsh Arora (UCLA), Yujia Peng (UCLA), Debleena Sengupta (UCLA), Lawrence Chen (UCLA), and Yuhe Gao (UCLA) for their help with conducting human studies and experiment setup, Prof. Devi Parikh (Georgia Tech), Prof. Dhruv Batra (Georgia Tech), Prof. Stefan Lee (OSU), for helpful discussions and useful feedback.



%% References
%% Usage of \citep is as follows:
%%   \citep{key}         ==>>  [#]
%%   \citep[chap. 2]{key} ==>> [#, chap. 2]
%%

%\bibliographystyle{elsarticle-harv}
%\bibliography{ref}

\end{document}

\begin{frontmatter}



\dochead{}
%% e.g. \dochead{17th International Conference on Dynamical Processes in Excited States of Solids}

\title{CX-ToM: Counterfactual Explanations with Theory-of-Mind for Enhancing Human Trust in Image Recognition Models}


\author{Arjun~R.~Akula*, %~\IEEEmembership{Member,~IEEE,}        
        Keze~Wang*, %~\IEEEmembership{Member,~IEEE,}
        Changsong~Liu, %~\IEEEmembership{Fellow,~IEEE,}
        Sari~Saba-Sadiya, %~\IEEEmembership{Member,~IEEE,}
        Hongjing~Lu, %~\IEEEmembership{Member,~IEEE,}
        Sinisa~Todorovic, %~\IEEEmembership{Fellow,~IEEE,}
        Joyce~Chai, %~\IEEEmembership{Fellow,~IEEE,}
        and~Song-Chun~Zhu%~\IEEEmembership{Fellow,~IEEE}% <-this % stops a space
%\IEEEcompsocitemizethanks{\IEEEcompsocthanksitem Arjun R. Akula: University of California, Los Angeles (UCLA), email: aakula@ucla.edu;\\ Keze Wang: UCLA, kezewang@gmail.com; \\ Changsong Liu: liucs.msu@gmail.com; \\ Sari~Saba-Sadiya: Michigan State University, sadiyasa@cse.msu.edu;\\ Hongjing Lu: UCLA, hongjing@ucla.edu; \\ Sinisa Todorovic: Oregon State University, sinisa@oregonstate.edu;\\ Joyce Chai: University of Michigan, Ann Arbor, jchai@cse.msu.edu;\\ Song-Chun Zhu: Beijing Institute for General Artificial Intelligence (BIGAI), Tsinghua University, Peking University, s.c.zhu@pku.edu.cn\protect}}
}

% \IEEEcompsocthanksitem * denotes equal contribution.\protect}% <-this % stops an unwanted space
% \thanks{Manuscript received Sep, 2020; revised Aug, 2021.}}

% \markboth{iScience Cell Press 2021}%
% {Shell \MakeLowercase{}}

\address{}

\end{frontmatter}

% \linenumbers






% \end{thebibliography}

\clearpage
\appendix
\section{Psychology Subject Pool Statistics}
\begin{figure*}[h]
\centering
  \includegraphics[width=\linewidth]{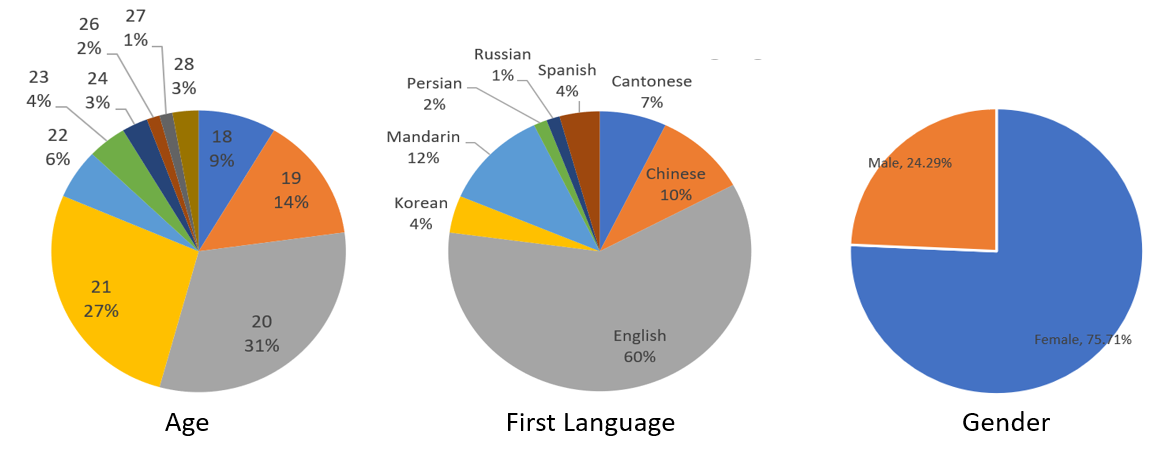}
  \caption{\small{Statistics (based on Age, First Language and Gender) of the 150 human subjects, from Psychology subject pool, participated in our study.}}~\label{fig:stats}
\end{figure*}

We provide more details on our human subject study here. Figure~\ref{fig:stats} shows the statistics (Age, First Language, Gender) of the human subjects, recruited from our institution's Psychology subject pool.

\section{Response Times}
\begin{figure*}[t]
\centering
  \includegraphics[width=\linewidth]{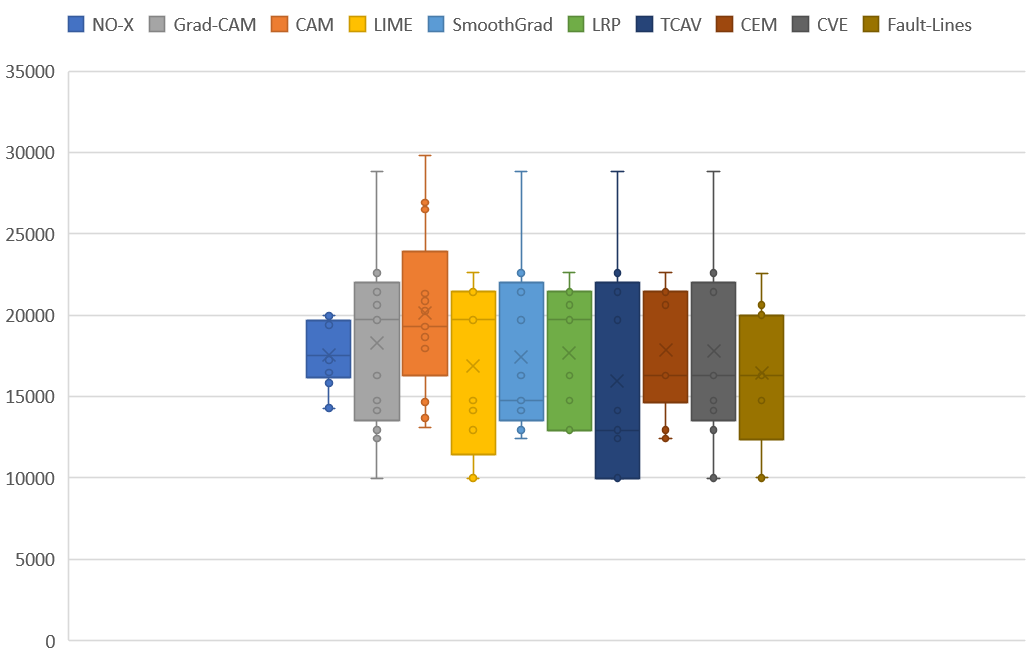}
  \caption{\small{\textbf{Response Times} (in milliseconds per question). Error bars denote standard errors of the means.}}~\label{fig:responsetime}
 
\end{figure*}

In addition to the qualitative and quantitative metrics discussed in the experiments section of the main paper, we also recorded the time taken by the human subject in answering the questions in the test phase. 
Figure~\ref{fig:responsetime} shows the average response times (in milliseconds per question) for each of the groups. 
We expected the participants in non-attention baselines such as TCAV, CEM and CVE, and in fault-line group to take less time for responding compared to the NO-X and attention baselines. Surprisingly, we find no significant difference in the response times across the groups.
% \begin{figure*}[t]
%   \centering
%   \mbox{
%     \subfigure{(a)\label{fig:responsetime}\includegraphics[width=6.5cm,height=4.5cm]{nips_app1.PNG}}\hfill
%     \subfigure{(b)\label{fig:subjective}\includegraphics[width=6.5cm,height=4.5cm]{nips_app2.PNG}}
%   }
%   \caption{\small{(a) \textbf{Response Times} (in milliseconds per question). (b) \textbf{Qualitative Reliance}. Error bars denote standard errors of the means.}}
% \vspace{-13pt}
% \end{figure*}

\section{More Examples of our Extracted Xconcepts}
\begin{figure*}[t]
\centering
  \includegraphics[width=0.7\linewidth]{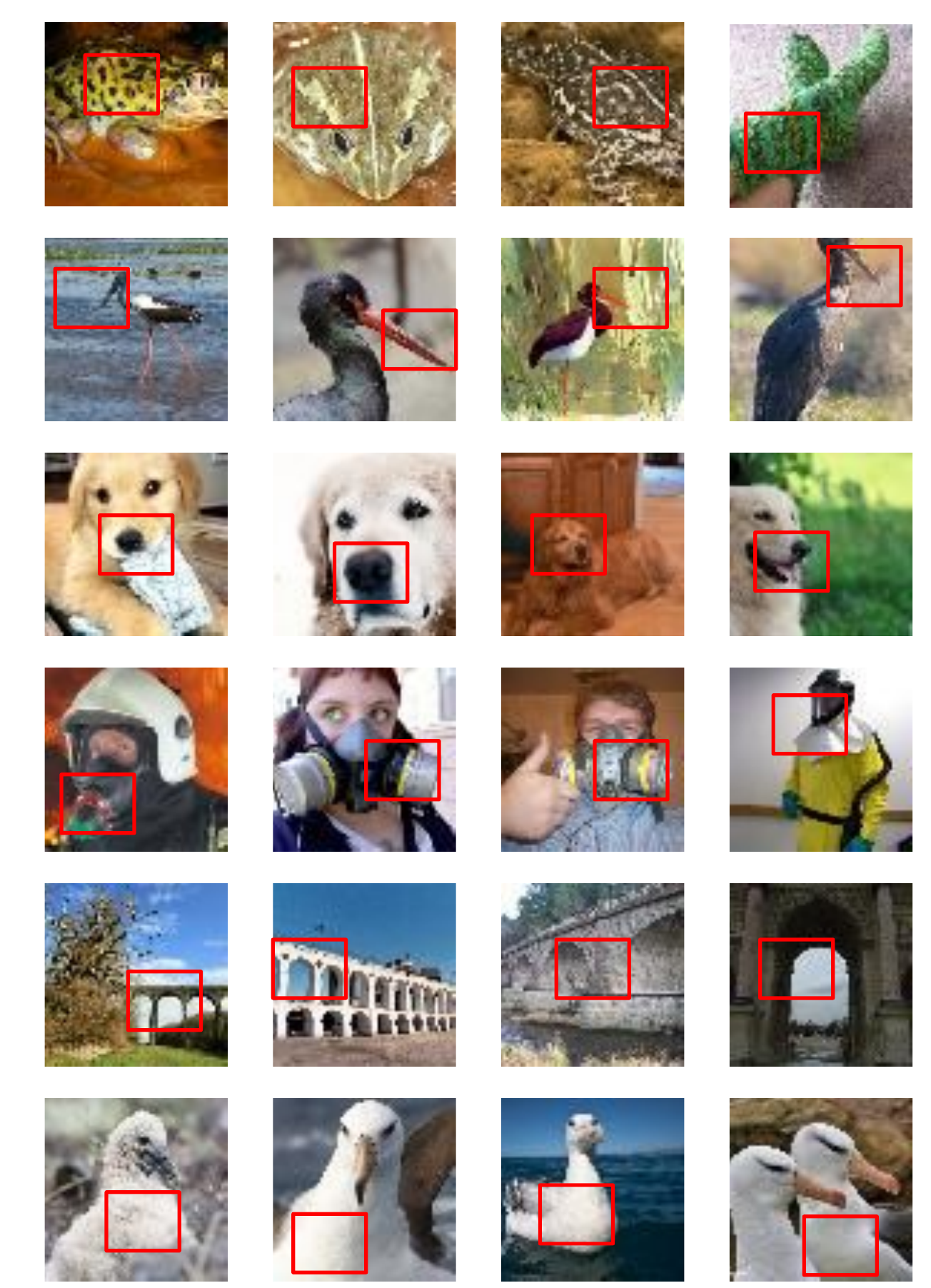}
  \caption{More examples for the Xconcepts generated in the section 3.2}~\label{fig:more_visualizations}
 
\end{figure*}

We provide more examples of the extracted xconcepts (along with the original image for clarity) in Figure~\ref{fig:more_visualizations}.

